\documentclass[10pt,twocolumn,letterpaper]{article}

\usepackage{cvpr}              % To produce the CAMERA-READY version
\definecolor{cvprblue}{rgb}{0.21,0.49,0.74}
\usepackage[pagebackref,breaklinks,colorlinks,allcolors=cvprblue]{hyperref}
\usepackage{array}
\usepackage{pifont}
\usepackage{amsthm}
\usepackage{fontawesome5}
\usepackage[table]{xcolor}
\def\paperID{16129} % *** Enter the Paper ID here
\def\confName{CVPR}
\def\confYear{2026}

\title{A Closed-Form Solution for Debiasing Vision-Language Models with Utility Guarantees Across Modalities and Tasks}

\author{Tangzheng Lian\\
King's College London\\
{\tt\small lian.tangzheng@kcl.ac.uk}
\and
Guanyu Hu\\
Queen Mary University of London\\
{\tt\small g.hu@qmul.ac.uk}
\and
Yijing Ren\\
King's College London\\
{\tt\small yijing.ren@kcl.ac.uk}
\and
Dimitrios Kollias\\
Queen Mary University of London\\
{\tt\small d.kollias@qmul.ac.uk}
\and
Oya Celiktutan\\
King's College London\\
{\tt\small oya.celiktutan@kcl.ac.uk}
}

\begin{document}
\maketitle
\begin{abstract}
While Vision-Language Models (VLMs) have achieved remarkable performance across diverse downstream tasks, recent studies have shown that they can inherit social biases from the training data and further propagate them into downstream applications. To address this issue, various debiasing approaches have been proposed, yet most of them aim to improve fairness without having a theoretical guarantee that the utility of the model is preserved. In this paper, we introduce a debiasing method that yields a \textbf{closed-form} solution in the cross-modal space, achieving Pareto-optimal fairness with \textbf{bounded utility losses}. Our method is \textbf{training-free}, requires \textbf{no annotated data}, and can jointly debias both visual and textual modalities across downstream tasks. Extensive experiments show that our method outperforms existing methods in debiasing VLMs across diverse fairness metrics and datasets for both group and \textbf{intersectional} fairness in downstream tasks such as zero-shot image classification, text-to-image retrieval, and text-to-image generation while preserving task performance. \textbf{Code} \url{https://github.com/Supltz/Debias\_VLM}.
\end{abstract}    
\section{Introduction}
\label{sec:intro}

% Transposed Table (✓ and ✗ only)
\begin{table*}[!ht]
\centering
\renewcommand{\arraystretch}{0.99} % row spacing
\setlength{\tabcolsep}{0.6pt}
\caption{Overview of recent debiasing methods in VLMs.}
{\footnotesize
\begin{tabular}{>{\centering\arraybackslash}m{3.8cm} 
                >{\centering\arraybackslash}m{0.8cm} 
                >{\centering\arraybackslash}m{0.8cm} 
                >{\centering\arraybackslash}m{1.1cm} 
                >{\centering\arraybackslash}m{1.3cm} 
                >{\centering\arraybackslash}m{1cm} 
                >{\centering\arraybackslash}m{1.3cm} 
                >{\centering\arraybackslash}m{1cm} 
                >{\centering\arraybackslash}m{1.4cm} 
                >{\centering\arraybackslash}m{1.1cm} 
                >{\centering\arraybackslash}m{1.3cm}
                >{\centering\arraybackslash}m{1cm}
                >{\centering\arraybackslash}m{0.5cm}}
\toprule
\textbf{Methods} & \textbf{SFID \cite{jung2024unified}} & \textbf{DeAR \cite{seth2023dear}} & \textbf{CLIP-clip \cite{wang-etal-2021-gender}} & \textbf{Fairer CLIP \cite{dehdashtian2024fairerclip}} & \textbf{PRISM \cite{molahasani2025prism}} & \textbf{PRISM-mini \cite{molahasani2025prism}} & \textbf{SANER \cite{hirota2025saner}} & \textbf{Biased Prompt \cite{chuang2023debiasing}} & \textbf{Robo Shot \cite{adila2023zeroshot}} & \textbf{Prompt Array \cite{berg-etal-2022-prompt}} & \textbf{JointV-L \cite{zhang2025joint}} &\textbf{\textcolor{red}{Ours}} \\
\midrule
\textbf{Training-Free} & \ding{56} & \ding{56} & \ding{51} & \ding{56} & \ding{56} & \ding{51} & \ding{56} & \ding{51} & \ding{51} & \ding{56} & \ding{56} &\textcolor{red}{\ding{51}} \\
\textbf{Data-Free} & \ding{56} & \ding{56} & \ding{56} & \ding{56} & \ding{51} & \ding{51} & \ding{51} & \ding{51} & \ding{51} & \ding{56} & \ding{56} & \textcolor{red}{\ding{51}} \\
\textbf{Debiasing Both Modalities} & \ding{51} & \ding{56} & \ding{51} & \ding{51} & \ding{51} & \ding{51} & \ding{56} & \ding{56} &  \ding{51} & \ding{51} & \ding{51} &\textcolor{red}{\ding{51}} \\
\rowcolor{gray!30} 
\textbf{Bounded Utility Loss} & \ding{56} & \ding{56} & \ding{56} & \ding{56} & \ding{56} & \ding{56} & \ding{56} & \ding{56} & \ding{56} & \ding{56} & \ding{56} & \textcolor{red}{\ding{51}} \\
\textbf{Zero-Shot Image Classification} & \ding{51} & \ding{56} & \ding{56} & \ding{51} & \ding{51} & \ding{51} & \ding{56} & \ding{51} & \ding{51} & \ding{56} & \ding{56} &\textcolor{red}{\ding{51}} \\
\textbf{Text-to-Image Retrieval} & \ding{51} & \ding{51} & \ding{51} & \ding{56} & \ding{56} & \ding{56} & \ding{51} & \ding{51} & \ding{56} & \ding{51} & \ding{51} &\textcolor{red}{\ding{51}} \\
\textbf{Text-to-Image Generation} & \ding{51} & \ding{56} & \ding{56} & \ding{56} & \ding{56} & \ding{56} & \ding{51} & \ding{51} & \ding{56} & \ding{56} & \ding{56} &\textcolor{red}{\ding{51}} \\
\rowcolor{gray!30} 
\textbf{Intersectional Fairness} & \ding{56} & \ding{56} & \ding{56} & \ding{56} & \ding{56} & \ding{56} & \ding{56} & \ding{56} & \ding{56} & \ding{56} & \ding{56} & \textcolor{red}{\ding{51}} \\
\bottomrule
\end{tabular}
}
\label{table1}
\end{table*}

VLMs such as CLIP (Contrastive Language–Image Pretraining) \cite{radford2021learning} have been applied and greatly improved performance across a wide range of downstream computer vision tasks by effectively bridging visual and linguistic semantics. This success largely stems from their training on massive collections of web-scraped image–text pairs \cite{schuhmann2021laion}, which enables strong generalization without the need for task-specific supervision. However, as shown in previous literature \cite{garcia2023uncurated, birhane2024dark}, these web-scraped contents inherently contain harmful stereotypes and spurious correlations arising from demographic imbalances. For example, in CLIP’s text embeddings, the concept “nurse” is unusually similar to “female”, while “doctor” is unusually similar to “male”, reflecting gender stereotypes that exist in online captions \cite{seth2023dear}. Similarly, web-scraped images of smiling faces often co-occur more frequently with female-looking faces than male-looking ones, even if there is no causal connection between gender and smiling expression \cite{lian2026feature, hu2025causalaffect}. Such biases or spurious correlations embedded within VLM representations can propagate into downstream applications if left unaddressed \cite{wang2023overwriting}. Ensuring algorithmic fairness in VLMs is therefore essential, particularly in human-centered computer vision tasks \cite{xu2021attention, lian2023supervised}, where discriminatory or stereotypical outputs can severely undermine public trust in AI systems \cite{thiebes2021trustworthy, hagendorff2020ethics, jobin2019global}.

However, debiasing VLMs has several key challenges:

(1) \textit{Multi-Modal Bias}. Bias in VLMs is encoded in both image and text embeddings by cross-modal alignment. Debiasing only one of them may not be sufficient \cite{zhang2025joint}.

(2) \textit{Training Cost.} Retraining VLMs on massive datasets from scratch is computationally impractical. Therefore, many debiasing methods freeze the image and text encoders and train additional debiasing networks on top of the embeddings \cite{seth2023dear, dehdashtian2024fairerclip, berg-etal-2022-prompt}, which not only require extra computational cost but also increase model complexity and inference cost by introducing extra network parameters to the VLMs.

(3) \textit{Sensitive Attribute Annotations.} Debiasing methods in VLMs typically require datasets annotated with sensitive attributes such as gender, race, or age \cite{jung2024unified, wang-etal-2021-gender}. Collecting such data is challenging due to privacy concerns and ethical constraints \cite{yang2020towards}. Moreover, perceived attribute annotations are often subjective and require extensive curation \cite{karkkainen2021fairface}. 

(4) \textit{Diverse Downstream Tasks.} VLMs are applied to a wide range of tasks, each with distinct fairness and utility metrics. Most existing debiasing methods focus on a single task, such as zero-shot image classification \cite{dehdashtian2024fairerclip, molahasani2025prism,adila2023zeroshot} or text-to-image retrieval \cite{seth2023dear,wang-etal-2021-gender, berg-etal-2022-prompt, zhang2025joint}, as it remains challenging to design a \textbf{\textit{unified}} debiasing approach with utility preservation across different downstream tasks \cite{hu2025fairmt}.

(5) \textit{Utility Preservation.} Debiasing VLMs often leads to a reduction in task performance \cite{seth2023dear, jung2024unified, berg-etal-2022-prompt}. Preserving downstream utility is challenging due to the inherent utility-fairness trade-off \cite{lian2025fair}. For example, in text-to-image generation, the input prompt may explicitly request generating male- or female-presenting images. Removing gender information can therefore degrade task performance \cite{hirota2025saner}.

Apart from these challenges, we also observe that although intersectional biases have been widely documented in VLMs \cite{Howard_2024_CVPR,hamidieh2024identifying}, prior debiasing methods have primarily focused on group fairness \cite{seth2023dear, dehdashtian2024fairerclip, molahasani2025prism, hirota2025saner, adila2023zeroshot, zhang2025joint, berg-etal-2022-prompt, jung2024unified,zhao2025bias}
, which only ensures parity across individual sensitive attributes. However, in real-world scenarios, individuals often belong to multiple sensitive groups simultaneously \cite{kearns2018preventing, molina2022bounding}. The concept of intersectional fairness remains largely underexplored in existing VLM debiasing studies. To bridge these gaps, our contributions are summarized as follows:  

(1) We propose a debiasing method that yields a \textbf{\textit{closed-form}} solution in the cross-modal space, achieving Pareto-optimal fairness with provably bounded utility losses.

(2) Our proposed method is \textbf{\textit{training-free}}, requires \textbf{\textit{no extra data}} and \textbf{\textit{sensitive attribute annotations}}, and can debias both visual and textual modalities across multiple tasks.

(3) Extensive experiments show that our method outperforms existing methods in debiasing VLMs across diverse fairness metrics and datasets for both group and \textbf{\textit{intersectional}} fairness in downstream tasks such as zero-shot image classification, text-to-image retrieval, and text-to-image generation, while preserving task performance.

\section{Related Work}
The presence of biases in VLMs has been extensively documented in prior research \cite{jiang-etal-2024-modscan, vo2025vision, weng2024images, ruggeri2023multi, hamidieh2024identifying, Howard_2024_CVPR}. In this section, we focus specifically on recent debiasing methods most relevant to our work, as summarized in Table \ref{table1}.

Many existing methods require both additional training of auxiliary networks and datasets annotated with sensitive attributes. For instance, DeAR \cite{seth2023dear} introduces a residual network to maximize the cross-entropy loss for predicting sensitive attributes. PromptArray \cite{berg-etal-2022-prompt} adopts a similar idea by prepending learnable embeddings and maximizing the prediction error of the adversary. FairerCLIP \cite{dehdashtian2024fairerclip} proposes a dependence metric to minimize the statistical dependence between embeddings and sensitive attributes, and JointV-L \cite{zhang2025joint} proposes to align bias across image and text modalities before applying counterfactual text to train a network.

SFID \cite{jung2024unified} and CLIP-clip \cite{wang-etal-2021-gender} both assume that certain embedding dimensions encode bias. SFID trains a lightweight Random Forest classifier to identify such biased dimensions and replaces them with neutral values, while CLIP-clip offers a training-free alternative by estimating mutual information and clipping those dimensions. However, both methods still require data with sensitive attribute annotations to locate such dimensions. PRISM (training-based) and its variant PRISM-mini (training-free) \cite{molahasani2025prism} debias embeddings via projection operations without extra annotations. However, similar to RoboShot \cite{adila2023zeroshot}, they only focus on zero-shot image classification tasks, leaving their applicability for other tasks such as text-to-image retrieval and generation unclear. SANER \cite{hirota2025saner} addresses bias in text-to-image retrieval and generation by proposing attribute neutralization. Although it does not rely on additional annotated data, it is not training-free. BiasedPrompt \cite{chuang2023debiasing} proposes two training-free methods: Orth-Proj \cite{chuang2023debiasing} and an extended version, Orth-Cali \cite{chuang2023debiasing}. However, both methods in BiasedPrompt and SANER focus solely on debiasing textual embeddings while ignoring biases in visual embeddings.

As shown in Table \ref{table1}, our method is the first to \textbf{\textit{simultaneously}} be training-free, data-free, and can debias both image and text modalities across all three downstream tasks. Regarding utility preservation, some approaches fail to explicitly address this aspect \cite{chuang2023debiasing}, while others, such as DeAR \cite{seth2023dear} and SANER \cite{hirota2025saner} attempt to preserve it through reconstruction losses. However, the ability to reconstruct embeddings via a trained network does not ensure that the original semantic utility or the cross-modal alignment is retained. Other methods \cite{jung2024unified, dehdashtian2024fairerclip, molahasani2025prism} claim utility preservation based on similar objectives and empirical results, often requiring extensive hyperparameter tuning. In contrast, we are the first to provide theoretical bounds on utility losses when debiasing VLMs. Moreover, prior methods primarily focus on group fairness. As noted in \cite{kearns2018preventing, molina2022bounding}, they can suffer from \textit{fairness gerrymandering}, where a model that is fair with respect to individual sensitive attributes but still be unfair at their intersections. To address this, we further extend the evaluation to intersectional fairness, capturing more complex and realistic scenarios when debiasing VLMs.

\section{Problem Formulation}

\noindent\textbf{Notation.} We denote the input image and text of VLMs as $I_{\mathrm{in}}$ and $T_{\mathrm{in}}$, and their corresponding encoders as $f_I(\cdot)$ and $f_T(\cdot)$. 
The resulting image and text embeddings are $\vec{e}_I = f_I(I_{\mathrm{in}})$ and $\vec{e}_T = f_T(T_{\mathrm{in}})$, respectively. 
We consider a sensitive attribute $S$, such as age, gender, or skin tone, or any intersection of them. 
Let $S$ contain $n \geq 2$ groups and define the set of attribute groups as $\mathcal{G} = \{ g_1, g_2, \dots, g_n \}$. 
For example, if $S$ represents \textit{gender}, then $\mathcal{G} = \{\text{male}, \text{female}\}$; 
if $S$ represents the intersection of \textit{gender} and \textit{race}, then 
$\mathcal{G} = \{ (\text{male}, \text{light}), (\text{female}, \text{dark}),\dots, (\text{female}, \text{light})\}$. We denote $\vec{u} \in \{\vec{u}_I, \vec{u}_T\}$ as the debiased embedding corresponding to the original embedding $\vec{e} \in \{\vec{e}_I, \vec{e}_T\}$, which we aim to find within the cross-modal space\footnote{The cross-modal space in VLMs is a unit hypersphere $\mathbb{S}^{d-1}$, where all embeddings are $\ell_2$-normalized and $d$ is the embedding dimension.} $\mathbb{S}^{d-1}$ for either the image or text modality (see details in Section \ref{4.2}).

\begin{figure}[t]
\hspace{0.2cm}
    \includegraphics[height=3.8cm, width=8cm]{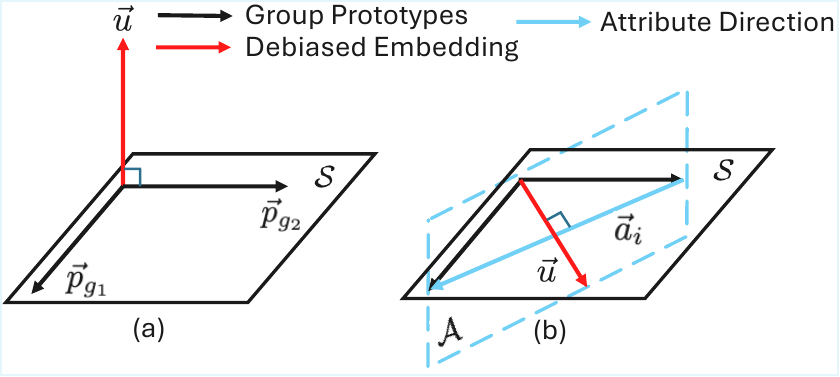} % Replace with your combined image file
    \caption{(a) Previous methods such as PRISM-mini \cite{molahasani2025prism} and Orth-Proj \cite{chuang2023debiasing} debias through an orthogonal projection to the subspace $\mathcal{S}$ spanned by the embeddings of group-explicit prompts (\textit{group prototypes}) such as ``a photo of a female/male doctor.'' However, the subspace $\mathcal{S}$ contains not only attribute-related information but also essential semantic content (e.g., ``doctor'') that we wish to preserve. (b) We instead debias through an orthogonal projection to the attribute subspace $\mathcal{A}$ so that $\vec{u}$ can effectively removes bias while maintaining content-related semantics.}

    \label{fig1}
\end{figure}

\subsection{Defining the Objectives}

\noindent\textbf{Utility}. We define utility from two perspectives. For downstream tasks that use a single encoder, such as text-to-image generation, we define \textbf{\textit{Self-Utility}} as the preservation of the original semantic content, measured by the cosine similarity between the debiased embedding $\vec{u}$ and the original embedding $\vec{e}$. We ensure self-utility by minimizing the self-utility loss of which we defined as $\ell_{self}=1-\langle \vec{u}, \vec{e}\, \rangle$ with a lower value indicating stronger semantic preservation. (2) For downstream tasks that utilize both encoders, such as zero-shot image classification, we define \textbf{\textit{Cross-Utility}} as the preservation of cross-modal alignment between the image–text pair before and after debiasing. We ensure cross-utility by minimizing the cross-utility loss denoted as $\ell_{cross}=\big| \langle \vec{u}_I, \vec{u}_T \rangle - \langle \vec{e}_I, \vec{e}_T \rangle \big|$ with a small value indicating that the image–text alignment is preserved.

%\footnote{This similarity-based fairness surrogate has been shown to improve multiple fairness metrics across downstream tasks effectively \cite{hirota2025saner, bolukbasi2016man}.} \footnote{Here, we choose group $g_1$ as the reference group, but any group could serve this role and result in the same attribute subspace. This is because the difference with respect to one reference group can be expressed as a linear combination of differences with respect to another, i.e., for any indices $i,j,k$, we have $p_{g_k} - p_{g_i} = (p_{g_k} - p_{g_j}) + (p_{g_j} - p_{g_i})$.}

\noindent\textbf{Fairness}. We formulate fairness as neutralization, which requires $\vec{u}$ to be equally similar\footnote{This similarity-based fairness surrogate has been shown to improve multiple fairness metrics across downstream tasks effectively \cite{hirota2025saner, bolukbasi2016man}.} to all embeddings of group-explicit prompts (\textit{group prototypes}): $\langle \vec{u}, \vec{p}_{g_1} \rangle = \langle \vec{u}, \vec{p}_{g_2} \rangle = \dots = \langle \vec{u}, \vec{p}_{g_n} \rangle$. This condition can be equivalently written as: $\langle \vec{u}, \vec{a}_{i} \rangle = 0, \;\forall i=2,\dots,n$, where $\vec{a}_{i} = \vec{p}_{g_i} - \vec{p}_{g_1}$ denotes the attribute direction between group prototypes. We thus define the attribute subspace as the span of these directions: $\mathcal{A}:= \mathrm{span}\{\vec{a}_2, \vec{a}_3, \dots, \vec{a}_n\}$. As shown in Fig \ref{fig1}, compared to \cite{chuang2023debiasing,molahasani2025prism} that project the embedding orthogonally onto the subspace $\mathcal{S}$, which may cause a loss of utility, this formulation ensures that $\vec{u}$ is orthogonal only to the attribute subspace, preserving the content-related semantics. 

\section{Methodology}

Our method consists of two main components. To construct the attribute subspace $\mathcal{A}$, for each group $g$, the first step builds textual prototypes $\vec{p}_g$ guided by a large language model (LLM). The second step then searches for the debiased embedding $\vec{u} \in \mathbb{S}^{d-1}$ while preserving the utilities. Please see \textbf{Appendix \textcolor{red}{A}} for the proofs of all propositions, lemmas, and theorems introduced in the methodology.

\begin{figure}[t]
\hspace{0.6cm}
    \includegraphics[height=6.4cm, width=7cm]{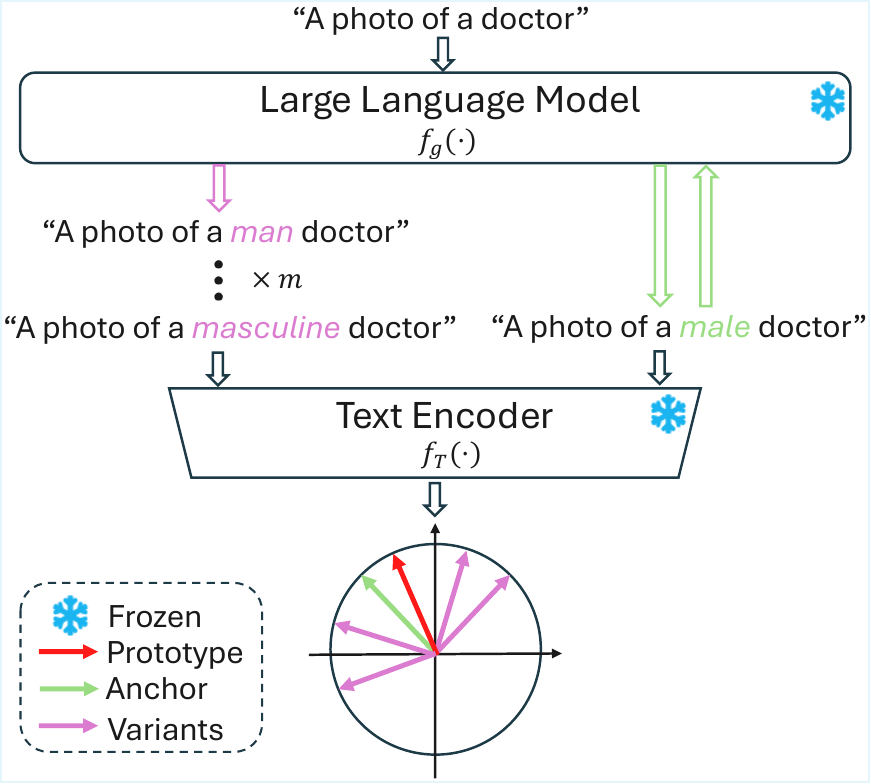} % Replace with your combined image file
    \caption{Illustration of the group prototype construction process.}

    \label{fig2}
\end{figure}

\subsection{LLM-Guided Group Prototype Construction}

Given an input prompt\footnote{We use the prompt template ``a photo of a/an \{sth\}'' only to align with prior debiasing studies \cite{hirota2025saner,jung2024unified}. This module is template-agnostic, and the input prompt could also be group-explicit (see \textbf{Appendix \textcolor{red}{B}}).} $T_{\mathrm{in}}$, e.g., ``a photo of a doctor,'' the goal of this module is to construct group prototypes conditioned on $T_{\mathrm{in}}$. As illustrated in Fig. \ref{fig2}, for each group $g$, we first pass $T_{\mathrm{in}}$ to a LLM, denoted as $f_{g}(\cdot)$, which performs two tasks within the same context window: (1) it inserts the group specifications while preserving the rest of the prompt content. For example, with $g =$ ``male,'' we may obtain $T_{g} = f_{g}(T_{\mathrm{in}}) =$ ``a photo of a male doctor.'' (2) it takes $T_{g}$ as input and generates multiple alternative phrasings that express $g$ while ensuring correct grammar and the rest of the text remains unchanged. For example, this may yield prompts such as ``a photo of a man doctor'' or ``a photo of a masculine doctor.'' We denote these variants as $\mathcal{T}_{g} = \{T_{g}^{(1)}, T_{g}^{(2)}, \dots, T_{g}^{(m)}\}$. We use GPT-5 as the LLM while other LLMs are also explored in Section \ref{4.6}. Please see more details of this module in \textbf{Appendix \textcolor{red}{B}}.

Previous works \cite{molahasani2025prism, chuang2023debiasing} directly use $T_{g}$ as the group prototype, overlooking that attribute groups are not linguistically monolithic. For instance, terms such as ``man,'' ``gentleman,'' and ``boy'' all convey different connotations of maleness. To address this, SANER \cite{hirota2025saner} constructs a corpus of group-specific words beforehand and inserts them into $T_{in}$. However, these words are not conditioned on $T_{in}$, which may lead to prompts that are semantically inconsistent with the input text. In contrast, we leverage the reasoning capability of LLMs to construct $\mathcal{T}_{g}$, ensuring better contextual alignment. We denote the embedding for the anchor prompt $T_{g}$ as $\vec{e}_{g} = f_T(T_{g})$. For the generated variants, we compute $\mathcal{E}_{g} = \{ \vec{e}_{g}^{\,(1)}, \vec{e}_{g}^{\,(2)}, \dots, \vec{e}_{g}^{\,(m)} \}$, where the embedding for each variant $ 
\vec{e}_{g}^{\,(i)} = f_T(T_{g}^{(i)})$.  To construct a representative group prototype, we seek an embedding on $\mathbb{S}^{d-1}$ that maximizes its cosine similarity with both $T_g$ and its generated variants, which corresponds to the spherical mean:
\[
\vec{p}_{g} = \frac{\vec{e}_{g} + \sum_{i=1}^m \vec{e}_{g}^{\,(i)}}{\big\| \vec{e}_{g} + \sum_{i=1}^m \vec{e}_{g}^{\,(i)} \big\|}.
\]

Therefore, for each group $g$, we set $\vec{p}_{g}$ as the group prototype. These prototypes are then used to construct the attribute space $\mathcal{A}$, to which $\vec{u}$ should be orthogonal.

\subsection{Searching the Debiased Embedding \texorpdfstring{\large $\vec{u}\in \mathbb{S}^{d-1}$}{u}}
\label{4.2}

\noindent\textbf{Preliminaries.} By the orthogonal decomposition theorem \cite{strang1993fundamental}, any embedding $\vec{u} \in \mathbb{S}^{d-1}$ can be decomposed as $\vec{u} = \vec{u}_{\mathcal{A}_\parallel} + \vec{u}_{\mathcal{A}_\perp}$, where $\vec{u}_{\mathcal{A}_\parallel} = P_{\mathcal{A}_\parallel}\vec{u}$ is the component parallel to $\mathcal{A}$ (i.e., the attribute leakage), and $\vec{u}_{\mathcal{A}_\perp} = P_{\mathcal{A}_\perp}\vec{u}$ is the component orthogonal to $\mathcal{A}$ (i.e., the neutral content). The projection operators are given by $P_{\mathcal{A}_\parallel} = A (A^\top A)^{-1} A^\top$ and $P_{\mathcal{A}_\perp} = I-P_{\mathcal{A}_\parallel}$, where $I$ is the identity matrix and $A = [\,\vec{a}_1, \vec{a_2}, \dots, \vec{a}_{r}\,] \in \mathbb{R}^{d \times r}$ is the basis matrix whose columns span $\mathcal{A}$ with $r = \mathrm{rank}(A) \leq n - 1 <<d$.

\noindent \textbf{Proposition 1.} The cross-utility loss $\ell_{cross}$ is upper bounded by $\sqrt{2\,\ell_{\text{self}}^{(I)}}+\sqrt{2\,\ell_{\text{self}}^{(T)}}$, where $\ell_{\text{self}}^{(I)}$ and $\ell_{\text{self}}^{(T)}$ are the self-utility loss of image and text embeddings, respectively.

By the decomposition, the fairness objective of finding $\vec{u}$ to be orthogonal to $\mathcal{A}$ can be formulated as minimizing attribute leakage ($||\vec{u}_{\mathcal{A}_\parallel}|| \to 0$). According to \textbf{Proposition 1}, the utility objective can be reduced to minimizing only the self-utility loss ($\ell_{self} \to 0$). Combining both objectives, we formulate a weighted-sum minimization problem as
\[
\min_{\vec{u}\in\mathbb{S}^{d-1}}G(\vec{u}):=w_1||\vec{u}_{\mathcal{A}_\parallel}||+w_2(1-\langle \vec{u}, \vec{e}\, \rangle),
\tag{1}
\]

\noindent where $w_1,w_2\geq0$ and $w_1+w_2=1$ denotes the preference.

\noindent \textbf{Lemma 1.} Every optimal solution $\vec{u}^{\,\star}$ of \textbf{Problem (1)} satisfies $\vec{u}^{\,\star}\in \text{span}\{\vec{e}_{\mathcal{A}_\parallel},\vec{e}_{\mathcal{A}_\perp}\}$, where $\vec{e}_{\mathcal{A}_\parallel}$ and $\vec{e}_{\mathcal{A}_\perp}$ denote the orthogonal components of $\vec{e}$ with respect to $\mathcal{A}$.

Solving \textbf{Problem (1)} is challenging because the search space is a unit hypersphere. By \textbf{Lemma 1}, the search space can be reduced to the 2D unit circle within $\text{span}\{\vec{e}_{\mathcal{A}_\parallel},\vec{e}_{\mathcal{A}_\perp}\}$. Hence, $\vec{u}$ can be represented as $\vec u=\alpha\,\vec e_{\mathcal A_\parallel}/||\vec e_{\mathcal A_\parallel}||+\beta\,\vec e_{\mathcal A_\perp}/||\vec e_{\mathcal A_\perp}||$, where $\alpha^2+\beta^2=1$. By this expression, we can scalarize the vector optimization in \textbf{Problem (1)} as
\[
\min_{\alpha^2+\beta^2=1}F(\alpha,\beta)
:=w_1|\alpha|+w_2(1-\alpha||\vec e_{\mathcal A_\parallel}||-\beta||\vec e_{\mathcal A_\perp}||).
\tag{2}
\]
\noindent\textbf{Proposition 2.} Debiasing is unnecessary when \textbf{Problem (2)} degenerates into 1D, i.e., $\|\vec{e}_{\mathcal{A}_\parallel}\|=0$ or $\|\vec{e}_{\mathcal{A}_\perp}\|=0$.

\begin{figure}[t]
\centering
\hspace{1.5cm}
    \includegraphics[height=3.8cm, width=4.6cm]{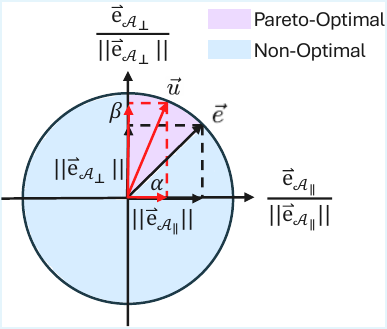} % Replace with your combined image file
    \caption{Illustration of the optimal solution space of $\vec{u}$.}
    \label{fig3}
\end{figure}

Therefore, we focus on the case where \textbf{Problem (2)} remains valid in 2D, i.e., $\|\vec{e}_{\mathcal{A}_\parallel}\|, \|\vec{e}_{\mathcal{A}_\perp}\| \in (0,1)$.

\noindent\textbf{Lemma 2.} The optimal solutions of \textbf{Problem (2)} lie in the first quadrant with $\alpha \in [0, \|\vec{e}_{\mathcal{A}_\parallel}\|]$, which fully characterizes the Pareto front where the attribute leakage and the self-utility loss are strictly increasing and decreasing. 

According to \textbf{Lemma 2}, $\vec{u}$ can be expressed solely using $\alpha$ as $\vec u=\alpha\,\vec e_{\mathcal A_\parallel}/||\vec e_{\mathcal A_\parallel}|| + \sqrt{1-\alpha^2}\,\vec e_{\mathcal A_\perp}/||\vec e_{\mathcal A_\perp}||$ and \textbf{Problem (2)} can then be rewritten as
\[
\min_{0\leq\alpha\leq||\vec e_{\mathcal A_\parallel}||}F(\alpha)
:=w_1L(\alpha)+w_2V(\alpha),
\tag{3}
\]
\noindent where $L(\alpha)=\alpha$ is the attribute leakage and $V(\alpha)=1-\alpha||\vec e_{\mathcal A_\parallel}||-\sqrt{1-\alpha^2}||\vec e_{\mathcal A_\perp}||$ is the self-utility loss.

As shown in Fig. \ref{fig3} and \textbf{Lemma 2}, we have two extreme cases of \textbf{Problem (3)}: (1) \textbf{\textit{Perfect Utility \& Worst Fairness}} $(\alpha = ||\vec e_{\mathcal A_\parallel}||)$: in this case, $\vec{u} = \vec{e}$, where the self-utility loss attains its minimum, $V(\alpha)=1-||\vec e_{\mathcal A_\parallel}||^2-||\vec e_{\mathcal A_\perp}||^2=0$ and the attribute leakage reaches its maximum, $L(\alpha)= ||\vec{e}_{\mathcal{A}_\parallel}||$. (2) \textbf{\textit{Perfect Fairness \& Worst Utility}} $(\alpha = 0)$: in this case, $\vec{u} = \vec{e}_{\mathcal{A}_\perp}/\|\vec{e}_{\mathcal{A}_\perp}\|$, where the attribute leakage achieves its minimum, $L(\alpha)= 0$ and the self-utility loss reaches its maximum, $V(\alpha)=1-||\vec e_{\mathcal A_\perp}||$. This extreme case corresponds exactly to the orthogonal projection operation adopted by previous methods \cite{chuang2023debiasing,molahasani2025prism, hirota2025saner}.

In contrast, we aim to solve \textbf{Problem (3)} to obtain a debiased embedding $\vec{u}^{\star}$ with $\alpha^{\star} \in (0, \|\vec{e}_{\mathcal{A}_\parallel}\|)$ such that it achieves Pareto-optimal fairness, while the self-utility loss $\ell_{self}$ is lower than $1 - \|\vec{e}_{\mathcal{A}_\perp}\|$. 
According to \textbf{Proposition 1}, the cross-utility violation $\ell_{cross}$ is also upper bounded by 
$\sqrt{2\,(1 - \|\vec{e}_{\mathcal{A}_\perp}^{\;(I)}\|)} + \sqrt{2\,(1 - \|\vec{e}_{\mathcal{A}_\perp}^{\;(T)}\|)}$. Since we aim for a \textit{task-agnostic} solution, we do not assume access to any data in the downstream tasks for task-specific tuning and searching $(w_1, w_2)$. Thus, we derive the optimal $\alpha^\star$ in \textbf{Problem (3)} that remains robust against any admissible $w_1$ and $w_2$ by minimizing the worst case via the Chebyshev scalarisation:
\[
\min_{0 \le \alpha \le \|\vec e_{\mathcal A_\parallel}\|}
\sup_{\substack{w_1,w_2\ge 0 \\ w_1+w_2=1}}
\Big\{
w_1\,L(\alpha)+ w_2\,V(\alpha) 
\Big\}
\tag{4}
\]
\noindent \textbf{Theorem 1.} For any admissible $w_1$ and $w_2$, \textbf{Problem (4)} admits a closed-form solution:
\[
\alpha^\star=\frac{E \ -\ \|\vec e_{\mathcal A_\perp}\|\sqrt{E^2-\|\vec e_{\mathcal A_\parallel}\|^2}}{\ E^2+\|\vec e_{\mathcal A_\perp}\|^2\ }.
\]

\noindent where $E:=\|\vec e_{\mathcal A_\parallel}\|+(1-\|\vec e_{\mathcal A_\perp}\|)/\|\vec e_{\mathcal A_\parallel}\|$ and the optimal debiased embedding is $\vec u^\star
=\sqrt{1-(\alpha^\star)^2} \vec e_{\mathcal A_\perp}/\|\vec e_{\mathcal A_\perp}\|+
\alpha^\star \vec e_{\mathcal A_\parallel}/\|\vec e_{\mathcal A_\parallel}\|$. The corresponding attribute leakage and self-utility loss are $L(\alpha^\star)=\alpha^\star$ and  $V(\alpha^\star)=(1-\|\vec e_{\mathcal A_\perp}\|)\alpha^\star/\|\vec e_{\mathcal A_\parallel}\|$, while a tighter upper bound on the cross-utility loss is given by
\[
\ell_{\text{cross}}\le \sqrt{\frac{2(1-\|\vec e_{\mathcal A_\perp}^{\;(I)}\|)\alpha^\star}{\|\vec e_{\mathcal A_\parallel}^{\;(I)}\|}}+\sqrt{\frac{2 (1-\|\vec e_{\mathcal A_\perp}^{\;(T)}\|)\alpha^\star}{\|\vec e_{\mathcal A_\parallel}^{\;(T)}\|}}.
\]
\section{Downstream Tasks and Their Evaluations}
%\footnote{Other metrics such as demographic parity and disparate impact do not account for the correctness of model predictions. In contrast, Equal Opportunity conditions on the true label, allowing fairness evaluation via the confusion matrix, which better supports decision-making \cite{hardt2016equality}.}
\textbf{Zero-Shot Image Classification.} This task jointly uses the image and text encoders to make predictions. Each candidate class is represented by a text prompt ``a photo of a/an \{class name\}'' and encoded using the text encoder, while the image encoder encodes the input image. The predicted class is the one whose text embedding has the highest cosine similarity with the encoded image. Bias in this setting manifests as disparities in performance across sensitive attribute groups for the same class. To quantify this, we adopt the widely used fairness metric \textit{Equal Opportunity (EO)} \cite{hardt2016equality} and measure its violation across groups and classes:
\[
\Delta_{\text{EO}}^{\text{Avg}} = \frac{1}{\mathcal{K}} \sum_{k=1}^{\mathcal{K}} 
\max_{g,g'} 
\left| 
p^{k}_{g} - 
p^{k}_{g'} 
\right|,
\]
\[
\Delta_{\text{EO}}^{\text{Max}} = 
\max_{k} 
\frac{2}{|\mathcal{G}|(|\mathcal{G}|-1)} \sum_{g,g'} 
\left| 
p^{k}_{g} - 
p^{k}_{g'} 
\right|.
\]

Here, $p^{k}_{g}=P(\hat{Y}=k | Y=k, \mathcal{G}=g)$ and $p^{k}_{g'}=P(\hat{Y}=k | Y=k, \mathcal{G}=g')$ denote the conditional true positive rates for unordered distinct groups $g$ and $g'$, respectively, while $\hat{Y}$ and $Y$ represent the predicted and true class. $\mathcal{K}$ and $\mathcal{G}$ denote the number of classes and attribute groups. Since both $\mathcal{K}$ and $\mathcal{G}$ can have more than two categories, following prior work \cite{jung2022learning}, we report two variants of EO violation in multi-class or multi-group scenarios: (1) $\Delta_{\text{EO}}^{\text{Avg}}$, which averages the violations across classes and takes the maximum across groups, and (2) $\Delta_{\text{EO}}^{\text{Max}}$, which averages within groups and then takes the maximum across classes. For utility, we use the (macro) F1 score as the evaluation metric.

\noindent\textbf{Text-to-Image Retrieval.} This task uses the joint matching capability of the image and text encoders in VLMs. Given an input prompt, the model retrieves images from a candidate dataset by ranking them based on the cosine similarity between image and text embeddings. Bias in this task occurs when the proportion of attribute groups in the retrieved results deviates from their original proportions in the candidate dataset. Following prior studies \cite{jung2024unified,hirota2025saner,chuang2023debiasing,seth2023dear,berg-etal-2022-prompt}, we adopt the \textit{MaxSkew@M} metric to evaluate fairness. Let $\gamma_g = N_g / N$, where $N_g$ denotes the number of images belonging to group $g$ in the candidate dataset, and $N$ is the total number of images in the candidate dataset. For each neutral prompt $t$, we define $\hat{\gamma}_g^t = M_g^t / M$, where $M_g^t$ is the number of images belonging to group $g$ in the top-$M$ retrieved images. The \textit{MaxSkew@M} metric is given by:
\[
\text{MS@M} = \frac{1}{|\mathcal{T}|} \sum_{t \in \mathcal{T}} \max_g \log\!\left(\frac{\hat{\gamma}_g^t}{\gamma_g}\right),
\]
where $\mathcal{T}$ denotes a set of text queries. For utility evaluation, we follow \cite{jung2024unified,wang-etal-2021-gender,berg-etal-2022-prompt} and use the \textit{Recall@K} (R@K) metric, which measures the probability that the ground-truth image appears among the top-$K$ retrieved results, with $K={5,10}$. Therefore, lower MS@M values indicate better fairness, while higher R@K values reflect stronger retrieval utility.

\noindent\textbf{Text-to-Image Generation.} Following \cite{chuang2023debiasing,hirota2025saner,jung2024unified}, we consider text-to-image generation as a representative downstream task where only one encoder (text encoder) of a VLM is used. Hence, we debias only the text embeddings to prevent them from propagating bias into the generative model. Bias in this task occurs when the distribution of attribute groups in generated images is not uniform for a neutral prompt. To quantify this, for each neutral prompt $t$, we adopt the \textit{Statistical Parity (SP)} metric \cite{hirota2025saner,teo2023measuring,chuang2023debiasing} to measure the deviation between the group distribution in generated images and the ideal uniform distribution:
\[
\mathrm{SP}_t = \sqrt{\sum_{g \in \mathcal{G}} \left( \frac{N_g^t}{N^t} - \frac{1}{|\mathcal{G}|} \right)^2},
\]
where $N_g^t$ is the number of generated images belonging to group $g$ and $N^t$ is the total number of generated images for each prompt. We use the prompt template ``A photo of a/an $o$,'' where $o$ is replaced with an occupation term. The full list of 34 occupations $\mathcal{T}_o$ is provided in \textbf{Appendix \textcolor{red}{C}}. For each prompt, we set $N^t=100$ with perceived gender (Male or Female) as the sensitive attribute. An unbiased model should yield $\mathrm{SP}_t$ close to zero, while higher values indicate greater bias. To compute $\mathrm{SP}_t$, attribute annotations of generated images are needed. Following prior works \cite{chuang2023debiasing,hirota2025saner}, we perform both quantitative and qualitative evaluations for fairness and utility in text-to-image generation.

\noindent\textbf{Fairness.}  For quantitative evaluation, following \cite{jung2024unified}, we use BLIP-2 \cite{li2023blip} to automatically annotate perceived gender in generated images. Since BLIP-2 may introduce annotation errors that distort fairness evaluation, qualitative evaluation is also performed. We have three independent annotators to label the top-5 prompts $\mathcal{T}_o^5 \subset \mathcal{T}_o$ showing the highest $\mathrm{SP}_t$ by BLIP-2. We then report two fairness metric: $\overline{\mathrm{SP}}_{\mathcal{T}_o}$ that average across $\mathcal{T}_o$ annotated by BLIP-2 and $\overline{\mathrm{SP}}_{5}$ that averages across $\mathcal{T}_o^5$ annotated by humans. 

\noindent\textbf{Utility.} Following \cite{hirota2025saner}, we evaluate utility by converting prompts in $\mathcal{T}_o$ into gender-explicit ones $\mathcal{T}_{e}$ (e.g., “a photo of a male doctor”). In this setting, the debiased model should generate images only corresponding to the specified gender. For quantitative evaluation, we use the CLIP score as an automated metric to measure average image–text alignment across $\mathcal{T}_{e}$. For qualitative evaluation, the same three annotators manually identify gender-mismatched images among the top-5 prompts in $\mathcal{T}_{e}$ with the lowest CLIP scores. The total number of mismatches is denoted as $N_I$, and we report the \textit{generation accuracy} computed as $\text{Acc}^\text{G} = (N_G - N_I)/N_G$, where $N_G = 500$ is the total number of generated images. Please refer to more details of the evaluation process in \textbf{Appendix \textcolor{red}{D}}.

\subsection{Datasets}
We focus on image datasets containing humans, as fairness is more well-defined in such contexts \cite{selbst}. For zero-shot image classification, we adopt two widely used fairness benchmark datasets: \textbf{CelebA} \cite{liu2015faceattributes} and \textbf{FACET} \cite{gustafson2023facet}. Following prior works \cite{chuang2023debiasing,molahasani2025prism,dehdashtian2024fairerclip,adila2023zeroshot}, we define a binary classification task that predicts hair color (blond vs. non-blond). To explore intersectional fairness, we extend the sensitive attribute to the intersection of perceived gender and age, resulting in four subgroups: Male–Old, Female–Old, Male–Young, and Female–Young. Since CelebA is a face-centric dataset with a binary label space, we additionally employ FACET to explore multi-class classification and full-body human images. The sensitive attribute in FACET is set as perceived gender (Male, Female) only.

For text-to-image retrieval, previous works \cite{hirota2025saner, chuang2023debiasing, seth2023dear} use datasets such as FairFace \cite{karkkainen2021fairface} and PATA \cite{seth2023dear} as candidate sets to compute MS@M. However, since these datasets contain only sensitive attribute annotations, they focus solely on fairness evaluation while neglecting task utility. To bridge this gap, we follow \cite{jung2024unified}, which adopts image–caption datasets as the candidate sets. Specifically, we use \textbf{Flickr30K} \cite{plummer2015flickr30k} and \textbf{COCO2017} \cite{lin2014microsoft} to compute both R@K and MS@M (M=1000). The sensitive attribute in Flickr30K is perceived gender (Male, Female), whereas in COCO2017 it is the intersection of perceived gender (Male, Female) and skin tone (Light, Dark). Please refer to \textbf{Appendix \textcolor{red}{E}} for a detailed description of all datasets used.

\begin{table*}[ht]
\centering
\renewcommand{\arraystretch}{1.3} % row spacing
\setlength{\tabcolsep}{2pt}
\caption{Experimental results for zero-shot image classification. (G) denotes perceived gender as the sensitive attribute, and (G$\times$A) denotes the intersection of perceived gender and age as the sensitive attribute. \textcolor{red}{$\uparrow$} indicates that higher values are better, while \textcolor{blue}{$\downarrow$} indicates that lower values are better. Methods that require or do not require attribute-annotated data are marked with \faCheck{} and \faTimes{}, respectively. Methods that involve or do not involve training are marked with \textcolor{orange}{\faFire{}} and \textcolor{cyan}{\faSnowflake{}}, respectively. Methods that debias only the text encoder are denoted by \textbf{\textcolor{red}{T}}, while methods that debias both modalities are marked with \textbf{\textcolor{red}{I\&T}}. The results of the debiasing methods with the best means are highlighted in bold with brackets, while the second-best are shown in brackets only. All values reported are scaled by 100 (percentages).}
\label{table2}
{\footnotesize
\begin{tabular}{l|c|cc|c|cc}
\hline
\textbf{Datasets} &
  \multicolumn{3}{c|}{CelebA} &
  \multicolumn{3}{c}{FACET} \\ \hline
\textbf{Evaluation Metric} &
  F1 \textcolor{red}{$\uparrow$} &
  $\Delta_{\text{EO}}^{\text{Avg}}$ (G$\times$A) \textcolor{blue}{$\downarrow$} &
  $\Delta_{\text{EO}}^{\text{Max}}$ (G$\times$A) \textcolor{blue}{$\downarrow$} &
  Macro F1 \textcolor{red}{$\uparrow$} &
  $\Delta_{\text{EO}}^{\text{Avg}}$ (G) \textcolor{blue}{$\downarrow$} &
  $\Delta_{\text{EO}}^{\text{Max}}$ (G) \textcolor{blue}{$\downarrow$} \\ \hline
\textbf{Baseline CLIP (ViT-L/14)} & 54.0$\pm$0.5 & 25.1$\pm$0.3 & 45.0$\pm$0.3 & 70.8$\pm$0.2 & 8.9$\pm$0.2 & 49.8$\pm$0.3 \\ \hline

\textbf{\faCheck $\,$ \textcolor{orange}{\faFire} \textcolor{red}{I\&T} SFID \cite{jung2024unified}} &
52.8$\pm$0.5 & 23.9$\pm$0.1 & 41.6$\pm$0.2 & 69.2 $\pm$0.3 & 9.3$\pm$0.4 & 49.8$\pm$0.2 \\ \hline

\textbf{\faCheck $\,$ \textcolor{orange}{\faFire} \textcolor{red}{I\&T} FairerCLIP \cite{dehdashtian2024fairerclip}} &
[53.1]$\pm$0.2 & 24.0$\pm$0.1 & 41.4$\pm$0.3 & [69.8]$\pm$0.2 & 9.2$\pm$0.2 & 50.1$\pm$0.4 \\ \hline

\textbf{\faTimes $\,$ \textcolor{orange}{\faFire} \textcolor{red}{I\&T} PRISM \cite{molahasani2025prism}} &
52.9$\pm$0.4 & 23.9$\pm$0.5 & 40.6$\pm$0.1 & 69.0$\pm$0.2 & \textbf{[8.1]}$\pm$0.3 & 48.3$\pm$0.3 \\ \hline

\textbf{\faTimes $\,$ \textcolor{cyan}{\faSnowflake} \textcolor{red}{I\&T} PRISM-mini \cite{molahasani2025prism}} &
50.4$\pm$0.1 & 24.2$\pm$0.6 & 40.8$\pm$0.3 & 69.2$\pm$0.4 & 8.6$\pm$0.2 & 49.2$\pm$0.5 \\ \hline

\textbf{\faTimes $\,$ \textcolor{cyan}{\faSnowflake} \textcolor{red}{I\&T} RoboShot \cite{adila2023zeroshot}} &
52.3$\pm$0.4 & \textbf{[23.3]}$\pm$0.2 & \textbf{[40.0]}$\pm$0.3 & 69.3$\pm$0.3 & 8.5$\pm$0.1 & \textbf{[47.3]}$\pm$0.5 \\ \hline

\textbf{\faTimes $\,$ \textcolor{cyan}{\faSnowflake} \textcolor{red}{T} Orth-Proj \cite{chuang2023debiasing}} &
49.3$\pm$0.6 & 26.0$\pm$0.2 & 42.1$\pm$0.4 & 68.6$\pm$0.3 & 9.0$\pm$0.5 & 50.2$\pm$0.1 \\ \hline

\textbf{\faTimes $\,$ \textcolor{cyan}{\faSnowflake} \textcolor{red}{T} Orth-Cali \cite{chuang2023debiasing}} &
50.1$\pm$0.2 & 26.1$\pm$0.4 & 42.4$\pm$0.6 & 69.3$\pm$0.1 & 9.2$\pm$0.3 & 50.4$\pm$0.5 \\ \hline

\rowcolor{gray!30}
\textbf{\faTimes $\,$ \textcolor{cyan}{\faSnowflake} \textcolor{red}{I\&T} Ours} &
\textbf{[56.5]}$\pm$0.5 & [23.6]$\pm$0.3 & [40.1]$\pm$0.3 & \textbf{[70.7]}$\pm$0.4 & [8.3]$\pm$0.1 & [47.5]$\pm$0.4 \\ \hline
\end{tabular}%
}
\end{table*}

\begin{table*}[ht]
\centering
\renewcommand{\arraystretch}{1.1} % row spacing
\setlength{\tabcolsep}{4pt}
\caption{Experimental results for text-to-image retrieval. The markers follow the same notation style as in Table \ref{table2}. (G$\times$ST) denotes the intersection of perceived gender and skin tone. All values reported are scaled by 100 (percentages).}
\label{table3}
{\footnotesize
\begin{tabular}{l|ccc|ccc}
\hline
\textbf{Datasets} &
  \multicolumn{3}{c|}{COCO2017} &
  \multicolumn{3}{c}{Flickr30K}  \\ \hline
\textbf{Evaluation Metric} &
  R@5 \textcolor{red}{$\uparrow$} &
  R@10 \textcolor{red}{$\uparrow$} &
  MS@1000 (G$\times$ST) \textcolor{blue}{$\downarrow$} &
  R@5 \textcolor{red}{$\uparrow$} &
  R@10 \textcolor{red}{$\uparrow$} &
  MS@1000 (G) \textcolor{blue}{$\downarrow$} \\ \hline

\textbf{Baseline CLIP (ViT-L/14)} &
83.8$\pm$0.3 & 90.1$\pm$0.4 & 13.4$\pm$0.5 &
91.0$\pm$0.6 & 95.4$\pm$0.4 & 20.3$\pm$0.3 \\ \hline

\textbf{\faCheck $\,$ \textcolor{orange}{\faFire} \textcolor{red}{I\&T} SFID \cite{jung2024unified}} &
[77.4]$\pm$0.4 & 86.6$\pm$0.2 & 13.2$\pm$0.3 &
86.8$\pm$0.3 & 90.7$\pm$0.5 & 13.6$\pm$0.5 \\ \hline

\textbf{\faCheck $\,$ \textcolor{orange}{\faFire} \textcolor{red}{I\&T} PromptArray \cite{berg-etal-2022-prompt}} &
77.2$\pm$0.6 & [86.7]$\pm$0.5 & 12.8$\pm$0.4 &
87.2$\pm$0.2 & 92.4$\pm$0.1 & 12.9$\pm$0.4 \\ \hline

\textbf{\faCheck $\,$ \textcolor{orange}{\faFire} \textcolor{red}{I\&T} FairerCLIP \cite{dehdashtian2024fairerclip}} &
76.8$\pm$0.3 & 85.4$\pm$0.4 & 10.2$\pm$0.6 &
[87.9]$\pm$0.5 & [92.5]$\pm$0.6 & 12.2$\pm$0.5 \\ \hline

\textbf{\faCheck $\,$ \textcolor{cyan}{\faSnowflake} \textcolor{red}{I\&T} CLIP-clip \cite{wang-etal-2021-gender}} &
76.1$\pm$0.2 & 85.2$\pm$0.3 & \textbf{[9.9]}$\pm$0.4 &
87.7$\pm$0.3 & 91.5$\pm$0.1 & \textbf{[11.7]}$\pm$0.5 \\ \hline

\textbf{\faTimes $\,$ \textcolor{cyan}{\faSnowflake} \textcolor{red}{T} Orth-Proj \cite{chuang2023debiasing}} &
74.5$\pm$0.4 & 83.6$\pm$0.3 & 13.6$\pm$0.5 &
84.4$\pm$0.2 & 88.9$\pm$0.1 & 14.1$\pm$0.2 \\ \hline

\textbf{\faTimes $\,$ \textcolor{cyan}{\faSnowflake} \textcolor{red}{T} Orth-Cali \cite{chuang2023debiasing}} &
75.2$\pm$0.5 & 84.9$\pm$0.4 & 13.8$\pm$0.2 &
85.3$\pm$0.4 & 90.7$\pm$0.3 & 14.3$\pm$0.6 \\ \hline

\rowcolor{gray!30}
\textbf{\faTimes $\,$ \textcolor{cyan}{\faSnowflake} \textcolor{red}{I\&T} Ours} &
\textbf{[81.1]}$\pm$0.3 & \textbf{[89.0]}$\pm$0.5 & [10.1]$\pm$0.4 &
\textbf{[90.4]}$\pm$0.5 & \textbf{[94.9]}$\pm$0.2 & [11.8]$\pm$0.3 \\ \hline

\end{tabular}%
}
\end{table*}

% \usepackage{graphicx}

\begin{figure*}[t]
  \centering
  \includegraphics[height=4.8cm, width=17.4cm]{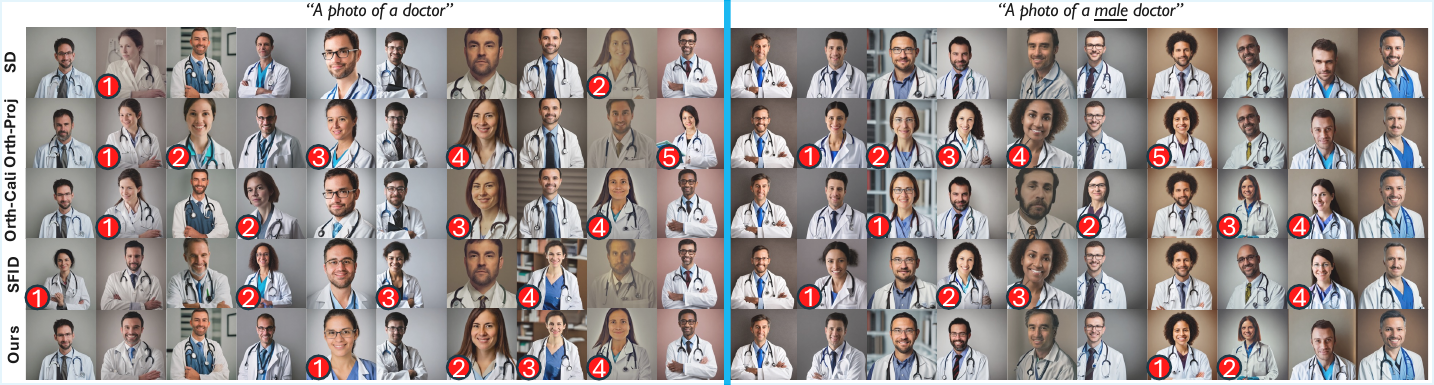}
\caption{Illustrative examples for $o=\text{``doctor"}$. We randomly sample ten generated images for each method. Female-looking samples are marked in \textcolor{red}{red} and numbered. On the left, a more balanced ratio of female- and male-looking samples indicates lower bias, while on the right, fewer female-looking samples reflect better preservation of self-utility (see \textbf{Appendix \textcolor{red}{G}} for more examples with other occupations).}
  \label{fig4}
\end{figure*}

\subsection{Experimental Results}

\noindent\textbf{Zero-Shot Image Classification and Text-to-Image Retrieval.} The VLMs we debias for these two tasks include CLIP \cite{radford2021learning} with different backbones (ViT-L/14 and ResNet50), as well as other models such as BLIP \cite{li2022blip}. Experimental results using CLIP (ViT-L/14) are presented in the main paper, while results for other models are provided in \textbf{Appendix \textcolor{red}{F}}. We compare our method with the \textbf{\textit{open-sourced}} debiasing approaches in Table \ref{table1} that have explicitly examined these tasks: \cite{jung2024unified, dehdashtian2024fairerclip, molahasani2025prism, adila2023zeroshot, chuang2023debiasing} for zero-shot image classification, and \cite{jung2024unified, dehdashtian2024fairerclip, chuang2023debiasing, berg-etal-2022-prompt, wang-etal-2021-gender} for text-to-image retrieval. The experimental results are presented in Tables \ref {table2} and \ref {table3}. Each experiment is conducted by bootstrapping the evaluation datasets, and we report the corresponding means and variances. Despite being both training-free and data-free, our method effectively debiases VLMs on these two tasks across multiple fairness metrics and datasets when compared with existing debiasing methods. Notably, it also maintains high task performance, whereas other methods exhibit notable performance drops.

Based on the empirical observations from our experiments, we further discuss the following key research questions (RQs) when designing debiasing methods for VLMs:

\noindent\textbf{RQ1: Are datasets annotated with sensitive attributes necessary for effective VLM debiasing?}

We find no systematic advantage of \faCheck$\,$methods over \faTimes$\,$methods in fairness evaluation across both tasks. In fact, \faCheck$\,$methods are highly sensitive to the domain of the annotated data. For instance, PromptArray, FairerCLIP, and SFID rely on face-centric datasets such as FairFace to train their debiasing networks, resulting in poor performance on full-body datasets with diverse scenes like FACET, COCO2017, and Flickr30K. As noted in \cite{zhang2025joint, seth2023dear}, to achieve robust debiasing, \faCheck$\,$methods require diverse annotated datasets covering multiple domains of images. However, collecting such large and diverse datasets with sensitive attribute labels is challenging in practice. In contrast, \faTimes$\,$methods overcome this limitation by exploring the generalizable representations encoded in CLIP, enabling effective debiasing without explicit attribute annotations.

\noindent\textbf{RQ2: Is training an additional mapping network necessary for effective VLM debiasing?}

We find no systematic advantage of \textcolor{orange}{\faFire} methods over \textcolor{cyan}{\faSnowflake} methods in fairness evaluation across both tasks. As illustrated in Fig. \ref{fig3}, we have proved that the region of Pareto-optimal debiased embeddings in the cross-modal space can be characterized through the orthogonal decomposition of the input embedding. Therefore, we argue that training an additional mapping network may complicate the problem.

\noindent\textbf{RQ3: Is it necessary to debias both image and text?}

For methods that are both \faTimes$\,$ and \textcolor{cyan}{\faSnowflake}, we observe that \textbf{\textcolor{red}{I\&T}} methods consistently outperform \textbf{\textcolor{red}{T}} methods on fairness metrics across both tasks. As noted in \cite{zhang2025joint}, this is because zero-shot image classification and text-to-image retrieval rely on both image and text embeddings, where biases are jointly encoded when aligning them through contrastive learning. This point is also confirmed by the ablation study of our method in Section \ref{4.6}.

\noindent\textbf{Text-to-Image Generation.} We debias the Stable Diffusion (SD) v2.1 model \cite{rombach2022high} using its default implementation. We compare our method with the \textbf{\textit{open-sourced}} debiasing methods \cite{chuang2023debiasing, jung2024unified} in Table \ref{table1}, which have explicitly examined this task. As shown in Fig. \ref{fig4}, our method generates balanced gender distributions for neutral prompts while effectively preserving model utility for gender-explicit ones with fewer wrong generations. The quantitative and qualitative results in Table \ref{table4} further demonstrate that our method achieves competitive debiasing performance while being the best in preserving utility compared with existing methods. Although Orth-Proj and Orth-Cali can effectively reduce bias, they do not explicitly address utility preservation, leading to lower CLIP scores and $\text{Acc}^{\text{G}}$.

% \usepackage{graphicx}
\begin{table}[t]
\centering
\renewcommand{\arraystretch}{1.3} % row spacing
\setlength{\tabcolsep}{3pt}
\caption{Experimental results for text-to-image generation. All values reported are scaled by 100 (percentages).}
\label{table4}
{\footnotesize
\begin{tabular}{l|c|c|c|c}
\hline
\textbf{Methods}   & $\overline{\mathrm{SP}}_{\mathcal{T}_o}$ \textcolor{blue}{$\downarrow$}& $\overline{\mathrm{SP}}_{5}$ \textcolor{blue}{$\downarrow$}& $\text{CLIP}_\text{score}$ (\%) \textcolor{red}{$\uparrow$} & $\text{Acc}^\text{G}$ (\%) \textcolor{red}{$\uparrow$} \\ \hline
\textbf{Baseline SD v2.1}        &  47.9  &  58.6  &  32.8  &  75.4   \\ \hline
\textbf{Orth-Proj \cite{chuang2023debiasing}}   &  \textbf{[39.6]}  & \textbf{[28.4]}  &  19.7  &  53.4  \\ \hline
\textbf{Orth-Cali \cite{chuang2023debiasing}}   &  41.5  & 29.2  &  20.7  &  56.6  \\ \hline
\textbf{SFID \cite{jung2024unified}} &   41.1 &  29.4  &  [22.6]  & [67.2]  \\ \hline
\rowcolor{gray!30} 
\textbf{Ours} &  [39.7]  &  [28.8]  &  \textbf{[24.2]}  &  \textbf{[74.6]} \\ \hline
\end{tabular}%
}
\end{table}

\begin{table}[t]
\centering
\renewcommand{\arraystretch}{1.3} % row spacing
\setlength{\tabcolsep}{3pt}
\caption{Experimental results of ablation studies and sensitivity analyses on different LLMs. $\vec p_a$ denotes the anchor embedding while $\vec p_m$ denotes the mean embedding across its variants.}
\label{table5}
{\footnotesize
\begin{tabular}{l|c|c}
\hline
\textbf{Datasets} &
  \multicolumn{1}{c|}{Flickr30K} &
  \multicolumn{1}{c}{CelebA} \\ \hline
\textbf{Metrics} &
  \multicolumn{1}{l|}{MS@1000 (G) \textcolor{blue}{$\downarrow$}} &
  $\Delta_{\text{EO}}^{\text{Max}}$ (G$\times$A) \textcolor{blue}{$\downarrow$} \\ \hline

\textbf{Baseline CLIP (ViT-L/14)} &
  20.3 $\pm$ 0.3 & 45.0 $\pm$ 0.3 \\ \hline

\textbf{debias with only $\vec p_a$} &
  13.4 $\pm$ 0.5 & 41.1 $\pm$ 0.4 \\ \hline

\textbf{debias with only $\vec p_m$} &
  14.1 $\pm$ 0.4 & 41.8 $\pm$ 0.5 \\ \hline

\textbf{debias with DeepSeek v3.2 \cite{liu2024deepseek}} &
  [12.0] $\pm$ 0.3 & \textbf{[40.1]} $\pm$ 0.4 \\ \hline

\textbf{debias with Gemini2.5 Pro \cite{comanici2025gemini}} &
  \textbf{[11.8]} $\pm$ 0.6 & [40.4] $\pm$ 0.2 \\ \hline

\textbf{debias with only $\vec{u}_I$} &
  13.4 $\pm$ 0.4 & 41.7 $\pm$ 0.5 \\ \hline

\textbf{debias with only $\vec{u}_T$} &
  13.3 $\pm$ 0.6 & 41.1 $\pm$ 0.3 \\ \hline

\rowcolor{gray!30}
\textbf{Ours} &
  \textbf{[11.8]} $\pm$ 0.3 & \textbf{[40.1]} $\pm$ 0.3 \\ \hline

\end{tabular}%
}
\end{table}

\subsubsection{Ablation Study \& Sensitivity Analyses}
\label{4.6}

We conduct ablation studies to evaluate the debiasing performance of our method and its sensitivity to different LLMs using the Flickr30K and CelebA as example datasets. As shown in Table \ref{table5}, using only the anchor embedding $\vec p_a$ or only the mean embedding $\vec p_m$ as the prototype results in poorer fairness metrics, validating the effectiveness of our proposed group prototype construction module. Moreover, applying debiasing to only one modality also leads to degraded debiasing performance on both datasets, confirming the necessity of jointly debiasing both modalities. Finally, our sensitivity analysis shows that different LLMs have minimal impact on debiasing performance. We attribute this to the simplicity of our prototype generation process, which can be easily handled by modern LLMs.

\section{Conclusion}
In this paper, we propose a \textbf{\textit{unified}} method that can debias VLMs while maintaining their utility across multiple downstream tasks. To this end, we derive a closed-form solution in the cross-modal space that attains Pareto-optimal fairness with bounded utility losses. Our method is \textbf{\textit{simple yet effective}}, can be easily plugged into any VLM, and it enjoys good properties such as being training-free, data-free, and capable of debiasing both image and text modalities. Extensive experiments show that our approach consistently outperforms existing VLM debiasing methods across multiple fairness metrics and datasets for both group and \textbf{\textit{intersectional}} fairness while effectively preserving utility.

\noindent\textbf{Limitations.} Our method focuses on debiasing within the cross-modal space constructed by the image and text encoders of VLMs. Thus, the utility guarantee is defined in this geometric space: it ensures semantic consistency within the VLM representation rather than directly guaranteeing performance on task-specific utility metrics such as F1 score or Recall@K. Extending the debiasing framework to image or text decoders for generative tasks is left for future work. In addition, due to limitations in existing fairness datasets, the attribute groups considered may not fully capture the full spectrum of sensitive attributes. Nonetheless, this does not undermine our methodological contribution, as our framework can accommodate any number of groups.

\noindent\textbf{Ethical Statement.} The sensitive attributes used in this work are either perceived or linguistically defined and do not correspond to the self-identification of any individual.

\section*{Acknowledgments}
Tangzheng Lian is fully funded by the PhD studentship in the Faculty of Natural, Mathematical and Engineering Sciences (NMES) at King's College London.
{
    \small
    \bibliographystyle{ieeenat_fullname}
    \bibliography{main}
}

 \clearpage

\setcounter{page}{1}

\onecolumn

\begin{center}
    {\Large \textbf{A Closed-Form Solution for Debiasing Vision-Language Models with Utility Guarantees Across Modalities and Tasks}} \\[0.5cm] % Supplementary title with line break
    {\Large Supplementary Material}
\end{center}
\vspace{0.5cm}

\noindent\textbf{\Large  Overview}\\

\begin{enumerate}
    \item \textbf{Appendix \textcolor{red}{A}}:  Proofs of the propositions, lemmas, and theorems.
    \item \textbf{Appendix \textcolor{red}{B}}: Details of the LLM-guided group prototype construction module.
    \item \textbf{Appendix \textcolor{red}{C}}:  The occupation list and its group-explicit variants used in text-to-image generation.
    \item \textbf{Appendix \textcolor{red}{D}}: The evaluation process in text-to-image generation.
    \item \textbf{Appendix \textcolor{red}{E}}: Detailed descriptions of the datasets used.
    \item \textbf{Appendix \textcolor{red}{F}}: Experimental results for CLIP (ResNet50) and BLIP.
    \item \textbf{Appendix \textcolor{red}{G}}: More illustrative examples for text-to-image generation.\\
\end{enumerate}

\noindent\textbf{\Large \textcolor{red}{A}. Proofs}\\

\noindent \textbf{Proof of Proposition 1.}
Denote the self-utility losses for the image and text as $\ell_{\text{self}}^{(I)}:=1-\langle \vec u_I,\vec e_I\rangle$ and $\ell_{\text{self}}^{(T)}:=1-\langle \vec u_T,\vec e_T\rangle$. For the cross-utility loss, we have
\[
\begin{aligned}
    \ell_{\text{cross}}=\big|\langle \vec u_I,\vec u_T\rangle-\langle \vec e_I,\vec e_T\rangle\big| &= \big|\langle \vec u_I,\vec u_T\rangle-\langle \vec e_I,\vec u_T\rangle
+\langle \vec e_I,\vec u_T\rangle-\langle \vec e_I,\vec e_T\rangle\big|\\&\stackrel{(1)}{\le} \big|\langle \vec u_I,\vec u_T\rangle-\langle \vec e_I,\vec u_T\rangle\big|+\big|\langle \vec e_I,\vec u_T\rangle-\langle \vec e_I,\vec e_T\rangle\big|\\&=\big|\langle \vec u_I-\vec e_I,\vec u_T\rangle\big|
+ \big|\langle \vec e_I,\vec u_T-\vec e_T\rangle\big|\\&\stackrel{(2)}{\le}\|\vec u_I-\vec e_I\|\;||\vec u_T||+||\vec e_I||\;\|\vec u_T-\vec e_T\|\\&=\|\vec u_I-\vec e_I\|+\|\vec u_T-\vec e_T\|.
\end{aligned}
\]

Here, $\stackrel{(1)}{\le}$ and $\stackrel{(2)}{\le}$ are because of the triangle inequality and Cauchy–Schwarz inequalities. For any unit vectors $\vec u,\vec e$, we have
\[
\|\vec u-\vec e\;\|^2
=\|\vec u\|^2+\|\vec e\,\|^2-2\langle \vec u,\vec e\,\rangle
=2\big(1-\langle \vec u,\vec e\,\rangle\big)
=2\,\ell_{\text{self}},
\]

Applying this to image and text gives
\[
\|\vec u_I-\vec e_I\|=\sqrt{2\,\ell_{\text{self}}^{(I)}},
\qquad
\|\vec u_T-\vec e_T\|=\sqrt{2\,\ell_{\text{self}}^{(T)}}.
\]

Then
\[
\ell_{\text{cross}}
\;\le\;
\|\vec u_I-\vec e_I\|+\|\vec u_T-\vec e_T\|
\;=\;
\sqrt{2\,\ell_{\text{self}}^{(I)}}+\sqrt{2\,\ell_{\text{self}}^{(T)}}
\]
\\
\qed\\

% \noindent\textbf{Proof of Proposition 2.} We solve the optimization problem
% \[
% \vec{p}_{g}=\max_{\vec{p} \in \mathbb{S}^{d-1}} \Big( \langle \vec{p}, \vec{e}_{g} \rangle + \sum_{i=1}^m \langle \vec{p}, \vec{e}_{g}^{\,(i)} \rangle \Big).
% \]
% For any $\vec{p} \in \mathbb{S}^{d-1}$, this can be rewritten as
% \[
% \max_{\vec{p} \in \mathbb{S}^{d-1}} \Big( \langle \vec{p}, \vec{e}_{g} \rangle + \sum_{i=1}^m \langle \vec{p}, \vec{e}_{g}^{\,(i)} \rangle \Big) = \max_{\vec{p} \in \mathbb{S}^{d-1}}  \Big\langle \vec{p}, \vec{e}_{g} + \sum_{i=1}^m  \vec{e}_{g}^{\,(i)} \Big\rangle
% \]

% By the Cauchy--Schwarz inequality,
% \[
% \Big\langle \vec{p}, \vec{e}_{g} + \sum_{i=1}^m  \vec{e}_{g}^{\,(i)} \Big\rangle \leq \|\vec{p}\,\| \, \|\vec{e}_{g} + \sum_{i=1}^m  \vec{e}_{g}^{\,(i)}\| = \|\vec{e}_{g} + \sum_{i=1}^m  \vec{e}_{g}^{\,(i)}\|,
% \]
% with equality \textit{iff} $\vec{p}$ is collinear with $\vec{e}_{g} + \sum_{i=1}^m  \vec{e}_{g}^{\,(i)}$. Hence, the unique embedding is obtained as
% \[
% \vec{p}_{g} = \frac{\vec{e}_{g} + \sum_{i=1}^m \vec{e}_{g}^{\,(i)}}{\big\| \vec{e}_{g} + \sum_{i=1}^m \vec{e}_{g}^{\,(i)} \big\|}.
% \]
% \\
% \qed
% \\

\noindent\textbf{Proof of Lemma 1.} The objective function in \textbf{Problem~(1)} can be rewritten as
\[
\begin{aligned}
G(\vec{u})&=w_1||\vec{u}_{\mathcal{A}_\parallel}||+w_2(1-\langle \vec{u}, \vec{e}\, \rangle) \\&= w_1||\vec{u}_{\mathcal{A}_\parallel}||+w_2\Big[1-\big(\langle \vec{u}_{\mathcal{A}_\parallel}, \vec{e}_{\mathcal{A}_\parallel}\, \rangle+\langle \vec{u}_{\mathcal{A}_\perp}, \vec{e}_{\mathcal{A}_\perp}\, \rangle\big)\Big]\\&\stackrel{(1)}{\geq}  w_1||\vec{u}_{\mathcal{A}_\parallel}||+w_2\Big[1-\big(|| \vec{u}_{\mathcal{A}_\parallel}||\, ||\vec{e}_{\mathcal{A}_\parallel}||+|| \vec{u}_{\mathcal{A}_\perp}||\,|| \vec{e}_{\mathcal{A}_\perp}||\big)\Big]
\end{aligned}
\]

Here, $\stackrel{(1)}{\geq}$ is because the Cauchy-Schwarz inequality $\langle \vec{u}_{\mathcal{A}_\parallel}, \vec{e}_{\mathcal{A}_\parallel}\, \rangle+\langle \vec{u}_{\mathcal{A}_\perp}, \vec{e}_{\mathcal{A}_\perp} \rangle \leq || \vec{u}_{\mathcal{A}_\parallel}||\, ||\vec{e}_{\mathcal{A}_\parallel}||+|| \vec{u}_{\mathcal{A}_\perp}||\,|| \vec{e}_{\mathcal{A}_\perp}||$ with the equality holds \textit{iff} both components of the orthogonal decompositions of $\vec{u}$ and $\vec{e}$ are collinear, i.e., $\vec{u}_{\mathcal{A}_\parallel}\parallel\vec{e}_{\mathcal{A}_\parallel}$ and $\vec{u}_{\mathcal{A}_\perp}\parallel\vec{e}_{\mathcal{A}_\perp}$. Since we are minimizing $G(\vec{u})$, any strict inequality in $\stackrel{(1)}{\geq}$ would mean that the right-hand collinear case produces a strictly smaller value of $G(\vec{u})$. Hence, no non-aligned $\vec{u}$ can be optimal. At optimality, $\stackrel{(1)}{\geq}$ must be tight and therefore $\vec{u}^{\,\star}\in \text{span}\{\vec{e}_{\mathcal{A}_\parallel},\vec{e}_{\mathcal{A}_\perp}\}$.
\\
\qed
\\

\noindent \textbf{Proof of Proposition 2.} There are two edge cases where the \textbf{Problem (2)} degenerates into 1D.\\

\noindent \textbf{Case A} ($\|\vec{e}_{\mathcal{A}_\parallel}\| = 0$ and $\|\vec{e}_{\mathcal{A}_\perp}\| = 1$).  
This means $\vec{e} = \vec{e}_{\mathcal{A}_\perp}$ is already fair. Then $F$ becomes
\[
F(\alpha, \beta)
= w_1 |\alpha| + w_2\big(1 - \beta\big).
\]
Here, no debiasing is needed, and we set $\alpha=0$ and  $\beta = \|\vec{e}_{\mathcal{A}_\perp}\|$ to fully preserve utility, which gives $\vec{u} = \vec{e} = \vec{e}_{\mathcal{A}_\perp}$ and $\ell_{\text{self}} = 1-||\vec{e}_{\mathcal{A}_\perp}||=0$.\\

\noindent \textbf{Case B} ($\|\vec{e}_{\mathcal{A}_\perp}\| = 0$ and $\|\vec{e}_{\mathcal{A}_\parallel}\| = 1$). This means $\vec{e} = \vec{e}_{\mathcal{A}_\parallel}$ contains only attribute information (e.g., prompts like ``a photo of a male'' or ``a photo of a female''). In this case, $F$ reduces to
\[
F(\alpha) = w_1 |\alpha| + w_2 (1 - \alpha).
\]
Since $\vec{e}$ reflects purely attribute-related semantics, debiasing is unnecessary. Therefore, we fully preserve its utility by setting $\alpha = ||\vec{e}_{\mathcal{A}_\parallel}||, \beta=0$, which yields $\vec{u} = \vec{e} = \vec{e}_{\mathcal{A}_\parallel}$ and $\ell_{\text{self}} = 1-||\vec{e}_{\mathcal{A}_\parallel}||=0$. 
\\
\qed
\\

\noindent\textbf{Proof of Lemma 2.} Assume $\beta < 0$ and define a sign-flipped feasible point $(\alpha, \tilde{\beta}) = (\alpha, -\beta)$, where $\tilde{\beta} > 0$.  
The objective difference between the two points is
\[
  \begin{aligned}
  F(\alpha,\tilde\beta)-F(\alpha,\beta)
  &= w_1|\alpha|+w_2(1-\alpha||\vec e_{\mathcal A_\parallel}||-\tilde\beta||\vec e_{\mathcal A_\perp}||) - \Big[w_1|\alpha|+w_2(1-\alpha||\vec e_{\mathcal A_\parallel}||-\beta||\vec e_{\mathcal A_\perp}||)\Big]\\&= w_2\big(-\tilde\beta||\vec e_{\mathcal A_\perp}||+\beta||\vec e_{\mathcal A_\perp}||\big)\
  \\&= w_2\big(-(-\beta)||\vec e_{\mathcal A_\perp}||+\beta||\vec e_{\mathcal A_\perp}||\big)\
  \\&= 2w_2\beta||\vec e_{\mathcal A_\perp}||
  \\&\stackrel{(1)}{\leq}0.
  \end{aligned}
\]

Here, $\stackrel{(1)}{\leq}$ because $w_2\geq 0$,$||\vec e_{\mathcal A_\perp}||> 0$ and $\beta<0$. Therefore, $F(\alpha, \tilde{\beta}) \le F(\alpha, \beta)$, meaning that a negative $\beta$ cannot yield a smaller objective value.  
Thus, the optimal solution must satisfy $\beta \ge 0$. A similar argument applies to $\alpha$.  
Assume $\alpha < 0$ and define $(\tilde{\alpha}, \beta) = (-\alpha, \beta)$, where $\tilde{\alpha} > 0$.  
Then,
\[
\begin{aligned}
F(\tilde\alpha,\beta)-F(\alpha,\beta) &= w_1|\tilde\alpha|+w_2(1-\tilde\alpha||\vec e_{\mathcal A_\parallel}||-\beta||\vec e_{\mathcal A_\perp}||) - \Big[w_1|\alpha|+w_2(1-\alpha||\vec e_{\mathcal A_\parallel}||-\beta||\vec e_{\mathcal A_\perp}||)\Big]\\
&\stackrel{(1)}{=}w_2\Big[-\tilde\alpha||\vec e_{\mathcal A_\parallel}||+\alpha||\vec e_{\mathcal A_\parallel}||\Big]\
\\&=w_2\Big[-(-\alpha)||\vec e_{\mathcal A_\parallel}||+\alpha||\vec e_{\mathcal A_\parallel}||\Big]\
\\&=2w_2\alpha||\vec e_{\mathcal A_\parallel}||
\\&\stackrel{(2)}{\leq} 0.
\end{aligned}
\]

Here,  $\stackrel{(1)}{=}$ because $|\tilde\alpha|=|-\alpha|=|\alpha|$ and $\stackrel{(2)}{\leq}$ because $w_2\geq 0$,$||\vec e_{\mathcal A_\parallel}||> 0$ and $\alpha<0$. Therefore, a negative $\alpha$ also cannot provide a smaller objective value, implying that the optimal solution must satisfy $\alpha \ge 0$.  
Hence, all optimal solutions of \textbf{Problem~(2)} lie in the first quadrant of the unit circle. We can then write the \textbf{Problem (2)} as

\[
\min_{0\leq\alpha\leq1}F(\alpha)
:=w_1L(\alpha)+w_2V(\alpha),
\]
\noindent where $L(\alpha)=\alpha$ is the attribute leakage and $V(\alpha)=1-\alpha||\vec e_{\mathcal A_\parallel}||-\sqrt{1-\alpha^2}||\vec e_{\mathcal A_\perp}||$ is the self-utility loss.

We further show that no optimal solution can lie in the interval $\|\vec e_{\mathcal A_\parallel}\| < \alpha \le 1$. Differentiating $F(\alpha)$ gives
\[
\frac{dF}{d\alpha}
= w_1\frac{dL}{d\alpha}+w_2\frac{dV}{d\alpha}
= w_1 + w_2\frac{dV}{d\alpha}.
\]

The derivative of $V(\alpha)$ is
\[
\frac{dV}{d\alpha}
= -\frac{d}{d\alpha}\Big(\alpha\,\|\vec e_{\mathcal A_\parallel}\|+\sqrt{1-\alpha^2}\,\|\vec e_{\mathcal A_\perp}\|\Big)
= \frac{\alpha}{\sqrt{1-\alpha^2}}\,\|\vec e_{\mathcal A_\perp}\|-\|\vec e_{\mathcal A_\parallel}\|.
\tag{$\alpha\neq1$}
\]

Setting $\frac{dV}{d\alpha}=0$ yields
\[
\frac{\alpha}{\sqrt{1-\alpha^2}}=\frac{\|\vec e_{\mathcal A_\parallel}\|}{\|\vec e_{\mathcal A_\perp}\|}= \frac{\|\vec e_{\mathcal A_\parallel}\|}{\sqrt{1-\|\vec e_{\mathcal A_\parallel}\|^2}}
\]

The unique solution to this equation is $\alpha=\|\vec e_{\mathcal A_\parallel}\|$.   To examine the sign of $\dfrac{dV}{d\alpha}$ around this point, define the function $h(x)=\dfrac{x}{\sqrt{1-x^2}}$ on $[0,1)$. Since
\[
h'(x)=\frac{1}{(1-x^2)^{3/2}}>0 \quad\text{for }x\in[0,1),
\]
$h(x)$ is strictly increasing on $[0,1)$. Consequently,
\[
\alpha>\|\vec e_{\mathcal A_\parallel}\| \ \Rightarrow\ h(\alpha) > h(\|\vec e_{\mathcal A_\parallel}\|) \Rightarrow\ \frac{\alpha}{\sqrt{1-\alpha^2}}>\frac{\|\vec e_{\mathcal A_\parallel}\|}{\sqrt{1-\|\vec e_{\mathcal A_\parallel}\|^2}}\ \Rightarrow\ \frac{dV}{d\alpha}>0,
\]
and conversely,
\[
\alpha<\|\vec e_{\mathcal A_\parallel}\| \ \Rightarrow\ h(\alpha) < h(\|\vec e_{\mathcal A_\parallel}\|) \Rightarrow\ \frac{\alpha}{\sqrt{1-\alpha^2}}<\frac{\|\vec e_{\mathcal A_\parallel}\|}{\sqrt{1-\|\vec e_{\mathcal A_\parallel}\|^2}}\ \Rightarrow\ \frac{dV}{d\alpha}<0,
\]

Since $w_1,w_2\ge0$ and $w_1+w_2=1$, when $\alpha>\|\vec e_{\mathcal A_\parallel}\|$, it follows that
\[
\frac{dF}{d\alpha}= w_1 + w_2\frac{dV}{d\alpha}
> 0
\]

Hence, $F(\alpha)$ is strictly increasing on $(\|\vec e_{\mathcal A_\parallel}\|,1)$ and cannot achieve an optimum in this interval. Next, we examine the boundary case $\alpha=1$.  
At $\alpha=1$, we have $F(1)=w_1+w_2(1-\|\vec e_{\mathcal A_\parallel}\|)$. At $\alpha=\|\vec e_{\mathcal A_\parallel}\|$, we have $F(\|\vec e_{\mathcal A_\parallel}\|)=w_1\|\vec e_{\mathcal A_\parallel}\|$. The difference between these two values is
\[
F(1)-F(\|\vec e_{\mathcal A_\parallel}\|)=w_1(1-\|\vec e_{\mathcal A_\perp}\|)+w_2(1-\|\vec e_{\mathcal A_\parallel}\|)\geq0,
\]

\noindent which confirms that $F(1)\geq F(\|\vec e_{\mathcal A_\parallel}\|)$.  
Thus, we can conclude that all optimal solutions must satisfy $0\leq\alpha\leq||\vec e_{\mathcal A_\parallel}||$.

Within this feasible range, we have $\frac{dL}{d\alpha}=1>0$ and $\frac{dV}{d\alpha}\le0$, implying that $L(\alpha)$ is strictly increasing while $V(\alpha)$ is strictly decreasing on $[0,\|\vec e_{\mathcal A_\parallel}\|]$.   For any two feasible points $\alpha_1<\alpha_2\le\|\vec e_{\mathcal A_\parallel}\|$, it follows that $L(\alpha_1)<L(\alpha_2)$ and $V(\alpha_1)>V(\alpha_2)$,  
indicating that reducing leakage necessarily increasing self-utility loss. Each $\alpha$ in this interval thus yields a distinct non-dominated pair $(L(\alpha),V(\alpha))$.  Therefore, the set $\alpha\in[0,\|\vec e_{\mathcal A_\parallel}\|]$ fully characterizes the Pareto front.
\\
\hfill\qed
\\

\noindent\textbf{Proof of Theorem 1.} For any scalars $(x,y)$ and weights $(w_1,w_2)\in\Delta:=\{w_1,w_2\ge 0,\ w_1+w_2=1\}$, we have
\[
\sup_{(w_1,w_2)\in\Delta}\big(w_1x+w_2y\big)=\max\{x,y\}.
\]
Therefore, the \textbf{Problem (3)} yields the following minimax formulation:
\[
\min_{0 \le \alpha \le \|\vec e_{\mathcal A_\parallel}\|}
\sup_{\substack{w_1,w_2\ge 0 \\ w_1+w_2=1}}
\Big\{
w_1\,L(\alpha)
+ w_2\,V(\alpha)
\Big\} = \min_{0 \le \alpha \le \|\vec e_{\mathcal A_\parallel}\|}\ \max\{ L(\alpha),\, V(\alpha)\}
\]

Since the two loss functions operate on different numerical scales over $\alpha \in [0, ||\vec e_{\mathcal A_\parallel}||]$, we normalize each to the interval $[0,1]$:
\[
\tilde L(\alpha)=\frac{\alpha}{\|\vec e_{\mathcal A_\parallel}\|},
\qquad
\tilde V(\alpha)=\frac{1 - \big(\sqrt{1-\alpha^2}\,\|\vec e_{\mathcal A_\perp}\| + \alpha\|\vec e_{\mathcal A_\parallel}\|\big)}{1-\|\vec e_{\mathcal A_\perp}\|}.
\]
This normalization ensures that $\tilde L(0)=0$, $\tilde L(||\vec e_{\mathcal A_\parallel}||)=1$,
and $\tilde V(0)=1$, $\tilde V(||\vec e_{\mathcal A_\parallel}||)=0$. Hence, \textbf{Problem (3)} becomes
\[
\min_{0 \le \alpha \le \|\vec e_{\mathcal A_\parallel}\|}\ \max\{\tilde L(\alpha),\,\tilde V(\alpha)\}
\]

By \textbf{Lemma 2}, over the interval $[0,||\vec e_{\mathcal A_\parallel}||]$,
$\tilde L(\alpha)$ is strictly increasing while $\tilde V(\alpha)$ is strictly decreasing.
Therefore, $\max\{\tilde L, \tilde V\}$ is minimized \textit{uniquely} at the point where they intersect:
\[
\tilde L(\alpha^\star)=\tilde V(\alpha^\star)
\quad\Longleftrightarrow\quad
\frac{\alpha^\star}{\|\vec e_{\mathcal A_\parallel}\|}
=
\frac{1-\big(\sqrt{1-(\alpha^\star)^2}\,\|\vec e_{\mathcal A_\perp}\|+\alpha^\star\|\vec e_{\mathcal A_\parallel}\|\big)}{1-\|\vec e_{\mathcal A_\perp}\|}.
\]

Rearranging, we obtain
\[
\|\vec e_{\mathcal A_\perp}\|\sqrt{1-(\alpha^\star)^2}
= 1 - \alpha^\star\Big(\|\vec e_{\mathcal A_\parallel}\| + \frac{1-\|\vec e_{\mathcal A_\perp}\|}{\|\vec e_{\mathcal A_\parallel}\|}\Big).
\]

For this equality to hold, the right-hand side must be positive, which provides an upper bound on $\alpha^\star$:
\[
\alpha^\star < \frac{1}{\displaystyle \|\vec e_{\mathcal A_\parallel}\|+\frac{1-\|\vec e_{\mathcal A_\perp}\|}{\|\vec e_{\mathcal A_\parallel}\|}}.
\tag{C2}
\]

Squaring both sides (as both are nonnegative) yields
\[
\|\vec e_{\mathcal A_\perp}\|^2(1-(\alpha^\star)^2)
=\Big(1 - \alpha^\star\Big(\|\vec e_{\mathcal A_\parallel}\|+\frac{1-\|\vec e_{\mathcal A_\perp}\|}{\|\vec e_{\mathcal A_\parallel}\|}\Big)\Big)^2.
\]

Expanding and rearranging leads to a quadratic in $\alpha^\star$:
\[
Q(\alpha^*):=\Big(\Big(\|\vec e_{\mathcal A_\parallel}\|+\frac{1-\|\vec e_{\mathcal A_\perp}\|}{\|\vec e_{\mathcal A_\parallel}\|}\Big)^2+\|\vec e_{\mathcal A_\perp}\|^2\Big)(\alpha^\star)^2
-2\Big(\|\vec e_{\mathcal A_\parallel}\|+\frac{1-\|\vec e_{\mathcal A_\perp}\|}{\|\vec e_{\mathcal A_\parallel}\|}\Big)\alpha^\star
+\|\vec e_{\mathcal A_\parallel}\|^2=0.
\]

The discriminant is
\[
\begin{aligned}
\Delta &= 4\Big[\big(\|\vec e_{\mathcal A_\parallel}\|+\frac{1-\|\vec e_{\mathcal A_\perp}\|}{\|\vec e_{\mathcal A_\parallel}\|}\big)^2-\Big(\Big(\|\vec e_{\mathcal A_\parallel}\|+\frac{1-\|\vec e_{\mathcal A_\perp}\|}{\|\vec e_{\mathcal A_\parallel}\|}\Big)^2+\|\vec e_{\mathcal A_\perp}\|^2\Big)\|\vec e_{\mathcal A_\parallel}\|^2\Big]\\&= 4\Big[\big(1-\|\vec e_{\mathcal A_\parallel}\|^2\big)\big(\|\vec e_{\mathcal A_\parallel}\|+\frac{1-\|\vec e_{\mathcal A_\perp}\|}{\|\vec e_{\mathcal A_\parallel}\|}\big)^2-\|\vec e_{\mathcal A_\perp}\|^2\|\vec e_{\mathcal A_\parallel}\|^2\Big]\\&=4\,\|\vec e_{\mathcal A_\perp}\|^2\,
\Big[\Big(\|\vec e_{\mathcal A_\parallel}\|+\frac{1-\|\vec e_{\mathcal A_\perp}\|}{\|\vec e_{\mathcal A_\parallel}\|}\Big)^2-\|\vec e_{\mathcal A_\parallel}\|^2\Big]\\&= 4\,\|\vec e_{\mathcal A_\perp}\|^2\,
\Big(\frac{(1-\|\vec e_{\mathcal A_\perp}\|)^2}{\|\vec e_{\mathcal A_\parallel}\|^2}
+2\big(1-\|\vec e_{\mathcal A_\perp}\|\big)\Big)\\&> 0
\end{aligned}
\]

Thus, there exist two real roots:
\[
\alpha_\pm
=\dfrac{
\Big(\|\vec e_{\mathcal A_\parallel}\|+\dfrac{1-\|\vec e_{\mathcal A_\perp}\|}{\|\vec e_{\mathcal A_\parallel}\|}\Big)
\ \pm\ 
\|\vec e_{\mathcal A_\perp}\|\sqrt{\Big(\|\vec e_{\mathcal A_\parallel}\|+\dfrac{1-\|\vec e_{\mathcal A_\perp}\|}{\|\vec e_{\mathcal A_\parallel}\|}\Big)^2-\|\vec e_{\mathcal A_\parallel}\|^2}
}{
\Big(\|\vec e_{\mathcal A_\parallel}\|+\dfrac{1-\|\vec e_{\mathcal A_\perp}\|}{\|\vec e_{\mathcal A_\parallel}\|}\Big)^2+\|\vec e_{\mathcal A_\perp}\|^2
}
\]

Since the solution is unique, we need to verify which root satisfies the feasibility conditions. \\

Let
\[
E:=\|\vec e_{\mathcal A_\parallel}\|+\frac{1-\|\vec e_{\mathcal A_\perp}\|}{\|\vec e_{\mathcal A_\parallel}\|}\,,
\qquad
\text{so that }\quad
\alpha_+=\frac{E \ +\ \|\vec e_{\mathcal A_\perp}\|\sqrt{E^2-\|\vec e_{\mathcal A_\parallel}\|^2}}{\ E^2+\|\vec e_{\mathcal A_\perp}\|^2\ }.
\]
The feasibility conditions are:
\[
\text{(C1)}\quad 0< \alpha^* < \|\vec e_{\mathcal A_\parallel}\|,
\qquad
\text{(C2)}\quad 
\alpha^* < \frac{1}{E}.
\]

We show that $\alpha_+$ violates at least one of these conditions.\\

\textbf{Case A: $\ E\ge 1$.}
Consider
\[
\alpha_+ - \frac{1}{E}
=
\frac{E+\|\vec e_{\mathcal A_\perp}\|\sqrt{E^2-\|\vec e_{\mathcal A_\parallel}\|^2}}{E^2+\|\vec e_{\mathcal A_\perp}\|^2}
-\frac{1}{E}
=
\frac{\ \|\vec e_{\mathcal A_\perp}\|\big(\,E\sqrt{E^2-\|\vec e_{\mathcal A_\parallel}\|^2}-\|\vec e_{\mathcal A_\perp}\|\,\big)}{\ E(E^2+\|\vec e_{\mathcal A_\perp}\|^2)\ }.
\]
The denominator is positive. For the numerator, since $E\ge 1$, we have
\[
\big(E\sqrt{E^2-\|\vec e_{\mathcal A_\parallel}\|^2}\big)^2-\|\vec e_{\mathcal A_\perp}\|^2
=(E^2-\|\vec e_{\mathcal A_\parallel}\|^2)E^2-\|\vec e_{\mathcal A_\perp}\|^2
=(E^2-1)\big(E^2+\|\vec e_{\mathcal A_\perp}\|^2\big)\ge 0,
\]
Hence,
\[
\big(E\sqrt{E^2-\|\vec e_{\mathcal A_\parallel}\|^2}\big)^2-\|\vec e_{\mathcal A_\perp}\|^2 = \Big(E\sqrt{E^2-\|\vec e_{\mathcal A_\parallel}\|^2}-\|\vec e_{\mathcal A_\perp}\|\Big)\Big(E\sqrt{E^2-\|\vec e_{\mathcal A_\parallel}\|^2}+\|\vec e_{\mathcal A_\perp}\|\Big) \ge0.
\]
Since 
$E\sqrt{E^2-\|\vec e_{\mathcal A_\parallel}\|^2}+\|\vec e_{\mathcal A_\perp}\|\ge 0$, we have
\[
E\sqrt{E^2-\|\vec e_{\mathcal A_\parallel}\|^2}-\|\vec e_{\mathcal A_\perp}\|\ge 0 \Longrightarrow \alpha_+\;-\;\frac{1}{E}\;\ge\;0.
\]

Thus, $\alpha_+$ violates (C2) when $E \ge 1$, making it infeasible.\\

\textbf{Case B: $\ E<1$.}
From $E=||\vec e_{\mathcal A_\parallel}||+\dfrac{1-||\vec e_{\mathcal A_\perp}||}{||\vec e_{\mathcal A_\parallel}||}<1$ with $||\vec e_{\mathcal A_\parallel}||^2+||\vec e_{\mathcal A_\perp}||^2=1$, we deduce
\[
1-\|\vec e_{\mathcal A_\perp}\|<\|\vec e_{\mathcal A_\parallel}\|(1-\|\vec e_{\mathcal A_\parallel}\|)\ \le\ \tfrac14
\quad\Longrightarrow\quad
\|\vec e_{\mathcal A_\perp}\|\ge\tfrac34\;>\;\tfrac12.
\]

Evaluating $Q(\alpha^*)$ at $\alpha^*=||\vec e_{\mathcal A_\parallel}||$ gives
\[
\begin{aligned}
Q\!\big(\|\vec e_{\mathcal A_\parallel}\|\big)
&=
\Big(\Big(\|\vec e_{\mathcal A_\parallel}\|+\frac{1-\|\vec e_{\mathcal A_\perp}\|}{\|\vec e_{\mathcal A_\parallel}\|}\Big)^2+\|\vec e_{\mathcal A_\perp}\|^2\Big)\|\vec e_{\mathcal A_\parallel}\|^2
-2\Big(\|\vec e_{\mathcal A_\parallel}\|+\frac{1-\|\vec e_{\mathcal A_\perp}\|}{\|\vec e_{\mathcal A_\parallel}\|}\Big)\|\vec e_{\mathcal A_\parallel}\|
+\|\vec e_{\mathcal A_\parallel}\|^2 \\
&=\Big(\|\vec e_{\mathcal A_\parallel}\|+\frac{1-\|\vec e_{\mathcal A_\perp}\|}{\|\vec e_{\mathcal A_\parallel}\|}\Big)^2 \|\vec e_{\mathcal A_\parallel}\|^2
+\|\vec e_{\mathcal A_\perp}\|^2\|\vec e_{\mathcal A_\parallel}\|^2
-2\Big(\|\vec e_{\mathcal A_\parallel}\|^2+1-\|\vec e_{\mathcal A_\perp}\|\Big)
+\|\vec e_{\mathcal A_\parallel}\|^2\\&= \Big[\|\vec e_{\mathcal A_\parallel}\|^4+2(1-\|\vec e_{\mathcal A_\perp}\|)\|\vec e_{\mathcal A_\parallel}\|^2+(1-\|\vec e_{\mathcal A_\perp}\|)^2\Big]
+\|\vec e_{\mathcal A_\perp}\|^2\|\vec e_{\mathcal A_\parallel}\|^2-2\Big(\|\vec e_{\mathcal A_\parallel}\|^2+1-\|\vec e_{\mathcal A_\perp}\|\Big)
+\|\vec e_{\mathcal A_\parallel}\|^2 \\&=\|\vec e_{\mathcal A_\parallel}\|^4
+\Big(2(1-\|\vec e_{\mathcal A_\perp}\|)+\|\vec e_{\mathcal A_\perp}\|^2-1\Big)\|\vec e_{\mathcal A_\parallel}\|^2
+\Big((1-\|\vec e_{\mathcal A_\perp}\|)^2-2(1-\|\vec e_{\mathcal A_\perp}\|)\Big)\\&= \|\vec e_{\mathcal A_\parallel}\|^4
+\big(\|\vec e_{\mathcal A_\perp}\|^2-2\|\vec e_{\mathcal A_\perp}\|+1\big)\|\vec e_{\mathcal A_\parallel}\|^2
+\big(\|\vec e_{\mathcal A_\perp}\|^2-1\big) \\&=\|\vec e_{\mathcal A_\parallel}\|^4
+\big((1-\|\vec e_{\mathcal A_\parallel}\|^2)-2\|\vec e_{\mathcal A_\perp}\|+1\big)\|\vec e_{\mathcal A_\parallel}\|^2
+\big((1-\|\vec e_{\mathcal A_\parallel}\|^2)-1\big) \\&=\|\vec e_{\mathcal A_\parallel}\|^4
+\big(2-2\|\vec e_{\mathcal A_\perp}\|-\|\vec e_{\mathcal A_\parallel}\|^2\big)\|\vec e_{\mathcal A_\parallel}\|^2
-\|\vec e_{\mathcal A_\parallel}\|^2 \\&=\|\vec e_{\mathcal A_\parallel}\|^4
+\big(2-2\|\vec e_{\mathcal A_\perp}\|\big)\|\vec e_{\mathcal A_\parallel}\|^2
-\|\vec e_{\mathcal A_\parallel}\|^4
-\|\vec e_{\mathcal A_\parallel}\|^2 \\
&=\|\vec e_{\mathcal A_\parallel}\|^2\big(1-2\|\vec e_{\mathcal A_\perp}\|\big)\\&<0 \qquad(\text{since }\|\vec e_{\mathcal A_\perp}\|>\tfrac12),
\end{aligned}
\]

Because the quadratic coefficient $E^2+||\vec e_{\mathcal A_\perp}||^2>0$, one root lies below $||\vec e_{\mathcal A_\parallel}||$ and the other lies strictly above it. The larger root corresponds to $\alpha_+$, so
\[
\alpha_+\;>\; \|\vec e_{\mathcal A_\parallel}\|,
\]
which violates (C1). Therefore, $\alpha_+$ is infeasible also when $E < 1$.

Combining \textbf{Cases A} and \textbf{B}, the $\alpha_+$ can never simultaneously satisfies both (C1) and (C2).

Therefore,
\[
\alpha^\star:=\alpha_-=
\dfrac{
\Big(\|\vec e_{\mathcal A_\parallel}\|+\dfrac{1-\|\vec e_{\mathcal A_\perp}\|}{\|\vec e_{\mathcal A_\parallel}\|}\Big)
-\|\vec e_{\mathcal A_\perp}\|\,
\sqrt{\Big(\|\vec e_{\mathcal A_\parallel}\|+\dfrac{1-\|\vec e_{\mathcal A_\perp}\|}{\|\vec e_{\mathcal A_\parallel}\|}\Big)^2-\|\vec e_{\mathcal A_\parallel}\|^2}
}{
\Big(\|\vec e_{\mathcal A_\parallel}\|+\dfrac{1-\|\vec e_{\mathcal A_\perp}\|}{\|\vec e_{\mathcal A_\parallel}\|}\Big)^2+\|\vec e_{\mathcal A_\perp}\|^2
}.
\]
The optimal debiased embedding is
\[
\;
\vec u^\star
=\sqrt{1-(\alpha^\star)^2}\,\frac{\vec e_{\mathcal A_\perp}}{\|\vec e_{\mathcal A_\perp}\|}
\ +\
\alpha^\star\,\frac{\vec e_{\mathcal A_\parallel}}{\|\vec e_{\mathcal A_\parallel}\|}\;.
\]

The attribute leakage and self-utility loss at $\vec u^\star$ are given by
\[
\; L(\alpha^\star)=\alpha^\star\;
\qquad\text{and}\qquad
\tilde L(\alpha^\star)=\tilde V(\alpha^\star)\ \Longrightarrow\ \frac{\alpha^\star}{\|\vec e_{\mathcal A_\parallel}\|}
=\frac{V(\alpha^\star)}{1-\|\vec e_{\mathcal A_\perp}\|} 
\; \ \Longrightarrow\ V(\alpha^\star)=(1-\|\vec e_{\mathcal A_\perp}\|)\,\frac{\alpha^\star}{\|\vec e_{\mathcal A_\parallel}\|}\; .
\]

Finally, by \textbf{Proposition 1}, we have a tighter upper bound on the cross-utility loss as:

\[
\ell_{\text{cross}}\le \sqrt{2(1-\|\vec e_{\mathcal A_\perp}^{\;(I)}\|)\,\frac{\alpha^\star}{\|\vec e_{\mathcal A_\parallel}^{\;(I)}\|}}+\sqrt{2 (1-\|\vec e_{\mathcal A_\perp}^{\;(T)}\|)\,\frac{\alpha^\star}{\|\vec e_{\mathcal A_\parallel}^{\;(T)}\|}}\le \sqrt{2\,(1 - \|\vec{e}_{\mathcal{A}_\perp}^{\;(I)}\|)} + \sqrt{2\,(1 - \|\vec{e}_{\mathcal{A}_\perp}^{\;(T)}\|)}
\]
\\
\qed
\\

\noindent\textbf{\Large \textcolor{red}{B}. Details of LLM}\\

\begin{figure*}[h]
  \centering
  \includegraphics[height=8.6cm, width=15cm]{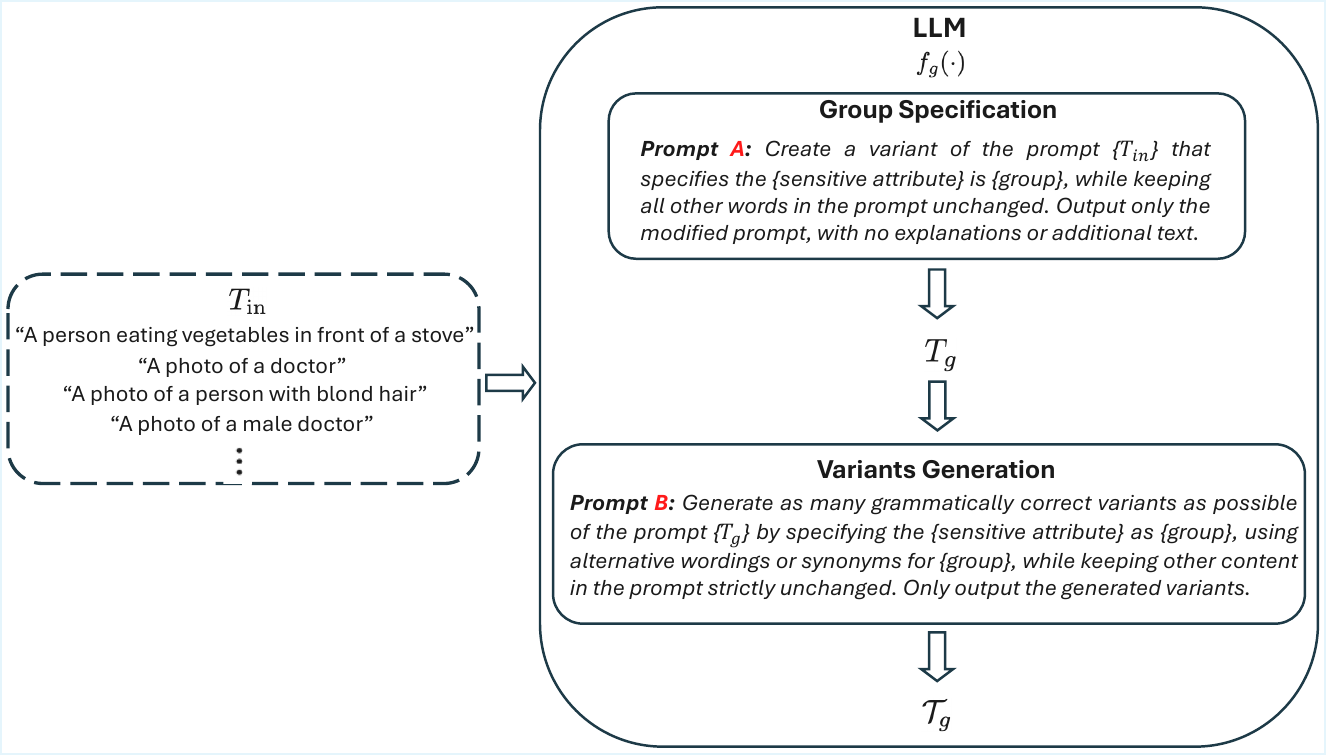}
\caption{An illustration of the prompts we use in an LLM to insert the group specification and generate the variants for each group $g$.}
\end{figure*}

As shown in Fig.~1, this module is template agnostic. The input prompt can take any form, whether neutral or group-specific, and all follow the same process. Once a sensitive attribute and its groups are defined, an LLM $f_{g}(\cdot)$ is used for each group to generate prompts. For each group, we collect the group-specific prompt $T_{g}$ and its variants $\mathcal{T}_{g}$, which are then fed into the text encoder of a VLM as described in Section~4.1 of the main paper. As shown in Fig. 2 and Fig. 3, we provide several examples of the LLM inputs and outputs.\\

\begin{figure*}[h]
  \centering
  \includegraphics[height=8.3cm, width=17.4cm]{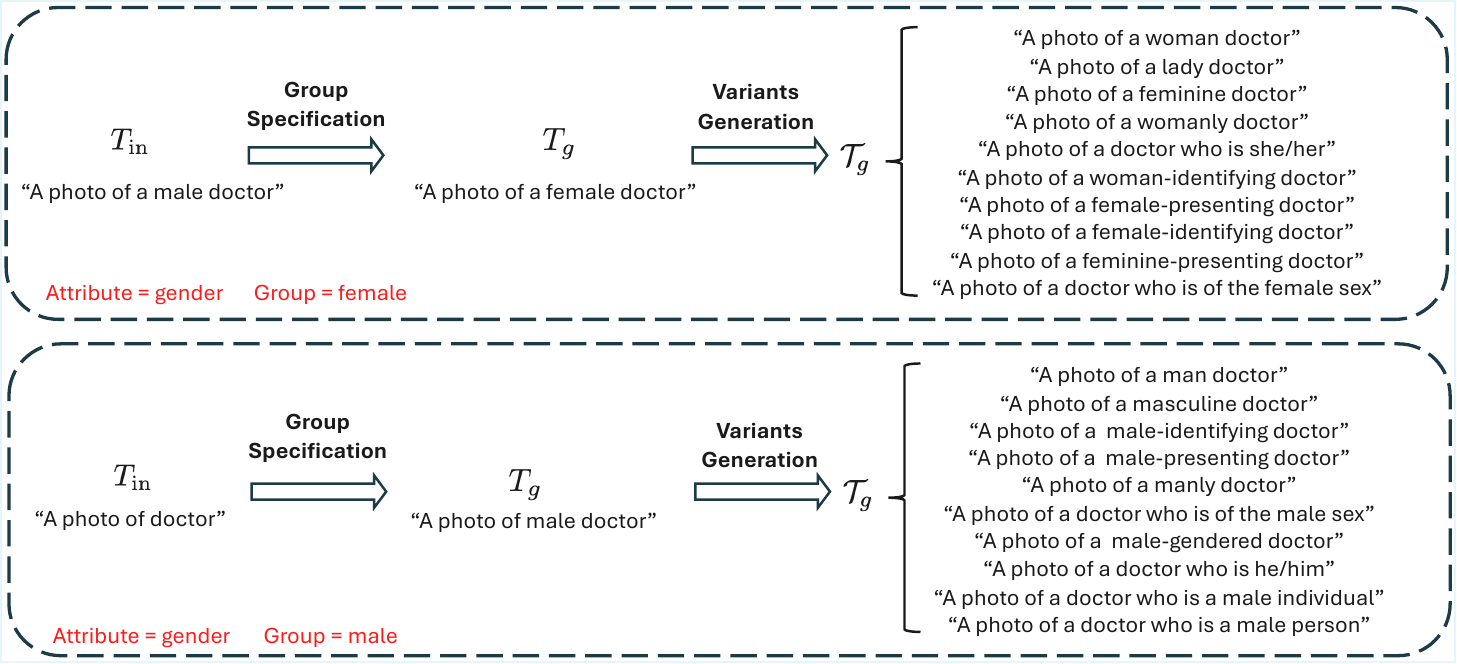}
\caption{An illustration of the LLM output when the input is group-specific/neutral. The sensitive attribute is gender, and the selected group is female/male.}
\end{figure*}

\vspace{0.4cm}

\begin{figure*}[h]
  \centering
  \includegraphics[height=7.5cm, width=17.4cm]{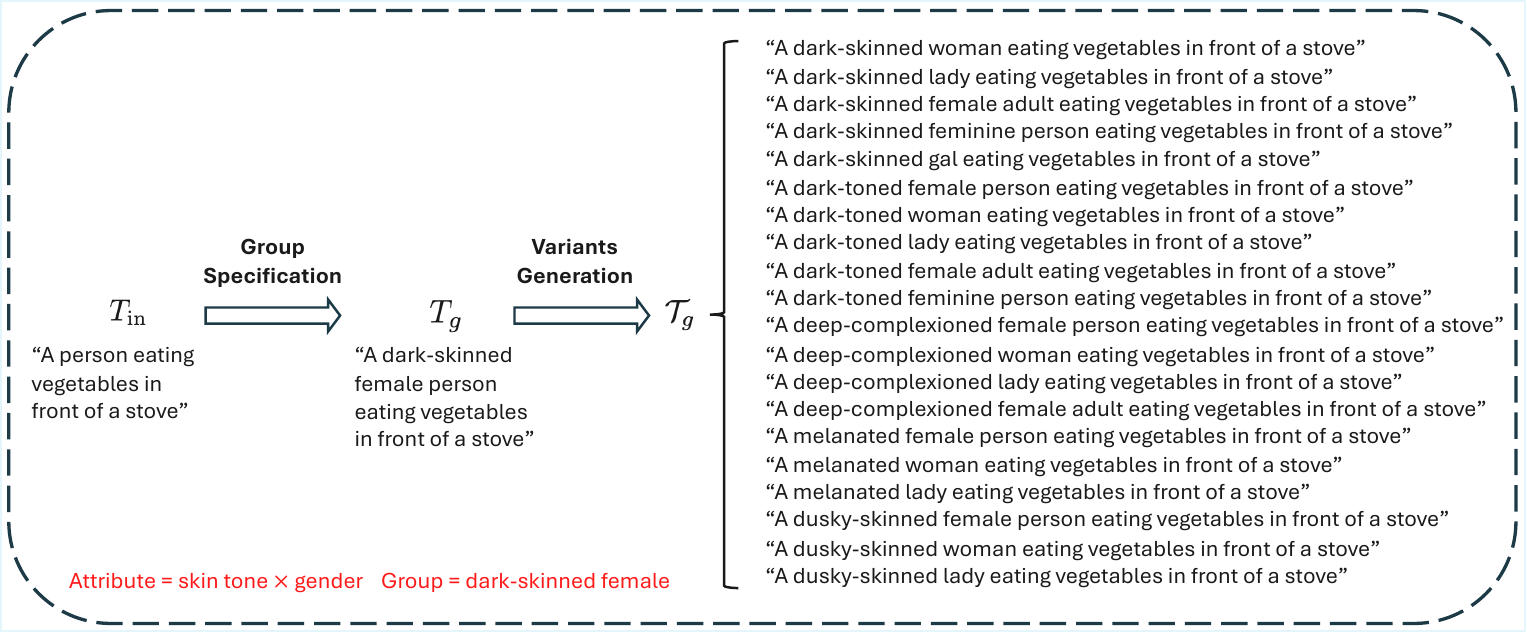}
\caption{An illustration of the LLM output when the input is neutral. The sensitive attribute is the intersection of skin tone and gender, and the selected group is dark-skinned female.}
\end{figure*}

\noindent\textbf{\Large \textcolor{red}{C}. Prompts in Text-to-Image Generation}\\

We use the following 34 occupations: 
\textit{doctor, aerospace engineer, computer programmer, electrical engineer, scientist, artist, designer, musician, painter, photographer, singer, writer, architect, civil engineer, engineer, software developer, childcare worker, coach, dentist, clerk, housekeeper, massage therapist, nurse, psychologist, social worker, teacher, professor, CEO, skateboarder, surfer, baseball player, football player, soccer player, tennis player}.

The occupation list is adapted from \cite{hirota2025saner, friedrich2023fair} with several modifications for clarity and redundancy reduction. Specifically, we remove \textit{maid} as it is inherently gendered, and delete \textit{dental assistant} and \textit{dental hygienist} due to their similarity to \textit{dentist}. We also exclude \textit{programmer} since \textit{computer programmer} is already included, and remove \textit{therapist} as it overlaps with \textit{massage therapist}. Additionally, \textit{author} is omitted because \textit{writer} serves the same purpose, and \textit{social assistant} is excluded due to the presence of \textit{social worker}. 

We then use these 34 occupations to create prompts following the template “a photo of a/an {occupation}.” As shown in Fig.~4, this results in 34 neutral prompts for fairness evaluation. We also construct 34 group-explicit prompts to evaluate utility. For each occupation, the group-explicit prompt specifies the gender group that the original model is biased toward. This choice is intentional, as many debiasing methods tend to overcorrect and fail to generate images for the originally dominant group once debiased. For example, the original SD model tends to generate male images for the occupation \textit{doctor}, while the debiased model balances the gender distribution but often struggles to generate male images when explicitly prompted, as it has been adjusted to produce more female representations.

\begin{figure*}[h]
  \centering
  \includegraphics[height=6cm, width=17.4cm]{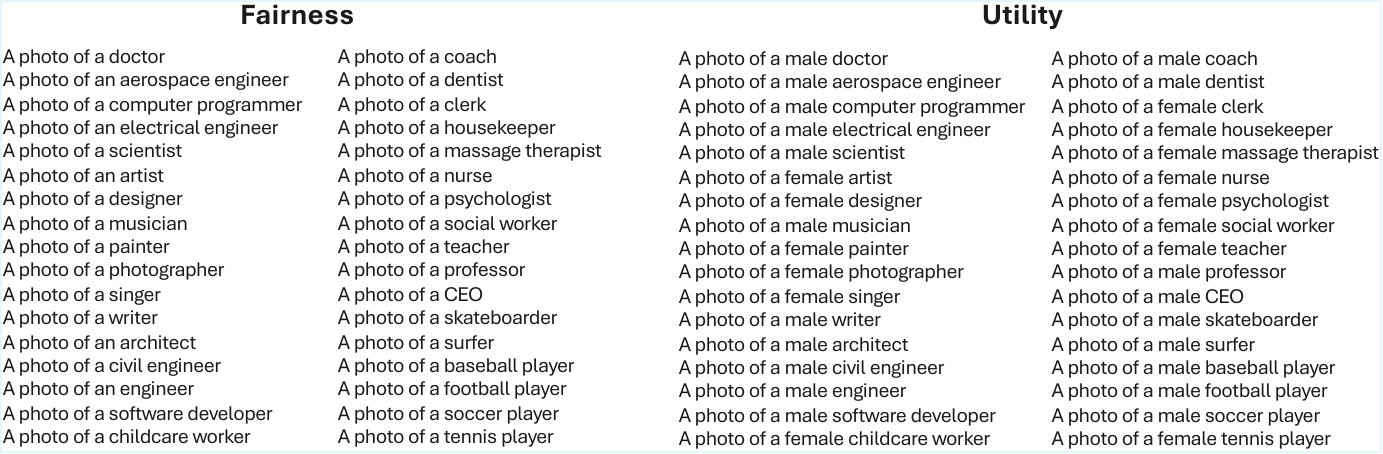}
\caption{An illustration of neutral prompts used for fairness evaluation and group-explicit prompts used for utility evaluation in the text-to-image generation task.}
\end{figure*}

% \begin{table}[ht]
% \centering
% \caption{The distribution of 52 person-related classes in the FACET dataset.}
% \begin{tabular}{l r l r l r l r}
% \toprule
% \textbf{Occupation} & \textbf{Count} &
% \textbf{Occupation} & \textbf{Count} &
% \textbf{Occupation} & \textbf{Count} &
% \textbf{Occupation} & \textbf{Count} \\
% \midrule
% astronaut & 286 & backpacker & 1612 & ballplayer & 1309 & bartender & 56 \\
% basketball\_player & 1668 & boatman & 2048 & carpenter & 223 & cheerleader & 399 \\
% climber & 455 & computer\_user & 1164 & craftsman & 1034 & dancer & 1397 \\
% disk\_jockey & 310 & doctor & 802 & drummer & 977 & electrician & 468 \\
% farmer & 1542 & fireman & 913 & flutist & 302 & gardener & 457 \\
% guard & 1361 & guitarist & 1180 & gymnast & 615 & hairdresser & 458 \\
% horseman & 735 & judge & 96 & laborer & 2540 & lawman & 4455 \\
% lifeguard & 511 & machinist & 354 & motorcyclist & 1367 & nurse & 1042 \\
% painter & 898 & patient & 884 & prayer & 798 & referee & 755 \\
% repairman & 1295 & reporter & 470 & retailer & 546 & runner & 638 \\
% sculptor & 213 & seller & 1178 & singer & 1286 & skateboarder & 990 \\
% soccer\_player & 1226 & soldier & 1457 & speaker & 1416 & student & 682 \\
% teacher & 192 & tennis\_player & 1661 & trumpeter & 498 & waiter & 332 \\
% \bottomrule
% \end{tabular}
% \label{tab:occupation_counts}
% \end{table}

\noindent\textbf{\Large \textcolor{red}{D}. Evaluations of Text-to-Image Generation}\\

For the \textbf{fairness evaluation}, we use 34 neutral prompts. For each prompt, we generate 100 images using five models: the original SD, Orth-Proj, Orth-Cali, SFID, and ours, resulting in a total of \textbf{17,000 generated images}. For quantitative evaluation, we query BLIP-2 with the question: ``What is the perceived gender of the person in this image?'' and compute $\overline{\mathrm{SP}}_{\mathcal{T}_o}$ based on BLIP-2’s annotations. For qualitative evaluation, we select the top-5 prompts $\mathcal{T}_o^5 \subset \mathcal{T}_o$ showing the highest $\mathrm{SP}_t$ scores under the original SD model: \textit{``A photo of a doctor,'' ``A photo of a CEO,'' ``A photo of a computer programmer,'' ``A photo of a designer,'' and ``A photo of a civil engineer.''} Each prompt has 100 generated images per model, totaling \textbf{2,500 images} for manual inspection. Three independent annotators answer the same question asked of BLIP-2 with three options: \textit{male}, \textit{female}, or \textit{unsure}. Images labeled as \textit{unsure} or with disagreement among annotators are regenerated until unanimous agreement is reached. As a result, all analyzed images have unanimously agreed perceived gender labels (male or female).

For the \textbf{utility evaluation}, we use 34 group-explicit prompts. Again, 100 images are generated per prompt for each of the five models, resulting in another \textbf{17,000 images}. For quantitative evaluation, we compute the CLIP score to measure image–text alignment and report the average score across all 34 prompts. For qualitative evaluation, we select the top-5 prompts with the lowest CLIP scores under the original SD model: \textit{``A photo of a male electrical engineer,'' ``A photo of a female photographer,'' ``A photo of a male professor,'' ``A photo of a male dentist,'' and ``A photo of a male football player.''} Each prompt produces 100 images per model, totaling \textbf{2,500 images} for manual inspection. The same three annotators identify images where the perceived gender does not match the one specified in the prompt. The number of mismatched images is denoted as $N_I$, and the \textit{generation accuracy} is computed as $\text{Acc}^\text{G} = (N_G - N_I)/N_G$, where $N_G = 500$ is the total number of images per prompt. Images labeled as \textit{unsure} or with non-unanimous annotations are regenerated until full agreement is reached, ensuring that all final images have unanimously agreed perceived gender labels.\\

\noindent\textbf{Ethical Statement.} We do not collect any personal information from the annotators, and all annotations are performed exclusively on AI-generated images. No personal data of real human subjects is involved in this study.\\

\noindent\textbf{\Large \textcolor{red}{E}. Datasets}\\

\noindent\textbf{CelebA.} The CelebA is a binary facial attribute public dataset widely used as a benchmark for fairness evaluation in computer vision. It consists of facial attribute images, and we use its original validation and test sets, totaling 39,829 images for both utility and fairness evaluation.\\

\noindent\textbf{FACET.} The FACET dataset is a publicly available benchmark created by Meta AI for evaluating fairness in computer vision. It contains approximately 32,000 images, each annotated with various attributes, including perceived gender, age, and skin tone. The original dataset provides bounding boxes for each person. By cropping each individual from the original images, we obtain 49,550 single-person images. For our experiments, we use perceived gender as the sensitive attribute. To maintain consistency with previous debiasing studies \cite{jung2024unified, chuang2023debiasing}, we exclude samples labeled as non-binary or unsure. The dataset also includes 52 occupation classes, which serve as the classification labels. To ensure statistically reliable results, we remove classes with fewer than 1000 samples, resulting in 26,834 images across 18 occupation categories. The retained occupations are:
\textit{lawman, laborer, boatman, basketball\_player, tennis\_player, backpacker, speaker, soldier, farmer, guard, dancer, singer, ballplayer, soccer\_player, repairman, guitarist, seller, motorcyclist.} \\

\noindent\textbf{Flickr30K.} Flickr30K is a public image–caption dataset containing 31,783 real-world images, each paired with up to five human-written descriptive sentences, resulting in a total of 158,915 image–text pairs. We first use YOLOv8 \cite{yolov5_2022} to select images containing only one person. For these selected images, we retain captions that include gender-explicit words from the list \textit{["man", "woman", "boy", "girl", "gentleman", "guy", "lady", "female", "male"]}. Captions containing any of \textit{["man", "boy", "gentleman", "guy", "male"]} are labeled as \textit{male}, while those with \textit{["woman", "girl", "lady", "female"]} are labeled as \textit{female}. After filtering, we obtain 3,079 image–text pairs for the retrieval task. To evaluate fairness, we further modify the captions by replacing gender-explicit words with the neutral term \textit{“person”}, and these neutralized captions are used as text queries.\\

\noindent\textbf{COCO2017.} The COCO2017 dataset is one of the most widely used image–caption datasets in computer vision, containing five human-written captions per image. We use a subset of the original training and validation sets provided by \cite{zhao2021understanding}, which includes explicit annotations for perceived gender and skin tone. These annotations were collected by multiple annotators using a more rigorous and reliable labeling protocol. In total, 28,316 images were annotated. We further exclude samples where either the perceived gender or skin tone is labeled as unsure, resulting in 1,368 image–text pairs. Similar to Flickr30K, to evaluate fairness, we modify the captions by replacing gender-explicit words with the neutral term \textit{“person”}, and use these neutralized captions as text queries.\\

\clearpage

\noindent\textbf{\Large \textcolor{red}{F}. Results of Other VLMs}\\

\begin{table*}[h]
\centering
\renewcommand{\arraystretch}{1.3} % row spacing
\setlength{\tabcolsep}{2pt}
\caption{Experimental results of CLIP (ResNet50) on the zero-shot image classification task.}
\label{table2}
{\small
\begin{tabular}{l|c|cc|c|cc}
\hline
\textbf{Datasets} &
  \multicolumn{3}{c|}{CelebA} &
  \multicolumn{3}{c}{FACET} \\ \hline
\textbf{Evaluation Metric} &
  F1 \textcolor{red}{$\uparrow$} &
  $\Delta_{\text{EO}}^{\text{Avg}}$ (G$\times$A) \textcolor{blue}{$\downarrow$} &
  $\Delta_{\text{EO}}^{\text{Max}}$ (G$\times$A) \textcolor{blue}{$\downarrow$} &
  Macro F1 \textcolor{red}{$\uparrow$} &
  $\Delta_{\text{EO}}^{\text{Avg}}$ (G) \textcolor{blue}{$\downarrow$} &
  $\Delta_{\text{EO}}^{\text{Max}}$ (G) \textcolor{blue}{$\downarrow$} \\ \hline

\textbf{Baseline CLIP (ResNet50)} & 
60.7$\pm$0.5 & 15.3$\pm$0.3 & 30.0$\pm$0.4 & 59.2$\pm$0.2 & 9.3$\pm$0.6 & 48.9$\pm$0.1 \\ \hline

\textbf{\faCheck $\,$ \textcolor{orange}{\faFire} \textcolor{red}{I\&T} SFID} &
56.4$\pm$0.2 & 14.6$\pm$0.5 & 24.9$\pm$0.3 & [50.8]$\pm$0.6 & 9.4$\pm$0.4 & 47.2$\pm$0.2 \\ \hline

\textbf{\faCheck $\,$ \textcolor{orange}{\faFire} \textcolor{red}{I\&T} FairerCLIP} &
[56.6]$\pm$0.3 & 14.2$\pm$0.1 & \textbf{[21.4]}$\pm$0.4 & 50.7$\pm$0.5 & 8.9$\pm$0.6 & 47.2$\pm$0.2 \\ \hline

\textbf{\faTimes $\,$ \textcolor{orange}{\faFire} \textcolor{red}{I\&T} PRISM} &
56.3$\pm$0.1 & 14.7$\pm$0.4 & [21.8]$\pm$0.3 & 50.5$\pm$0.6 & \textbf{[8.4]}$\pm$0.5 & 47.6$\pm$0.2 \\ \hline

\textbf{\faTimes $\,$ \textcolor{cyan}{\faSnowflake} \textcolor{red}{I\&T} PRISM-mini} &
56.0$\pm$0.6 & 13.7$\pm$0.3 & 24.9$\pm$0.1 & 49.2$\pm$0.4 & 8.9$\pm$0.5 & 46.6$\pm$0.2 \\ \hline

\textbf{\faTimes $\,$ \textcolor{cyan}{\faSnowflake} \textcolor{red}{I\&T} RoboShot} &
55.8$\pm$0.4 & \textbf{[10.9]}$\pm$0.2 & \textbf{[21.4]}$\pm$0.3 & 50.7$\pm$0.1 & 9.0$\pm$0.6 & \textbf{[45.6]}$\pm$0.5 \\ \hline

\textbf{\faTimes $\,$ \textcolor{cyan}{\faSnowflake} \textcolor{red}{T} Orth-Proj} &
56.4$\pm$0.3 & 14.6$\pm$0.5 & 24.4$\pm$0.2 & 47.8$\pm$0.6 & 8.8$\pm$0.4 & 47.9$\pm$0.1 \\ \hline

\textbf{\faTimes $\,$ \textcolor{cyan}{\faSnowflake} \textcolor{red}{T} Orth-Cali} &
55.6$\pm$0.2 & 14.5$\pm$0.4 & 24.5$\pm$0.3 & 49.6$\pm$0.6 & 8.9$\pm$0.1 & 47.6$\pm$0.5 \\ \hline

\rowcolor{gray!30}
\textbf{\faTimes $\,$ \textcolor{cyan}{\faSnowflake} \textcolor{red}{I\&T} Ours} &
\textbf{[58.5]}$\pm$0.4 & [11.5]$\pm$0.3 & [21.8]$\pm$0.2 & \textbf{[58.6]}$\pm$0.1 & [8.5]$\pm$0.2 & [45.8]$\pm$0.5 \\ \hline

\end{tabular}%
}
\end{table*}

\begin{table*}[h]
\centering
\renewcommand{\arraystretch}{1.1}
\setlength{\tabcolsep}{4pt}
\caption{Experimental results of CLIP (ResNet50) on the text-to-image retrieval task.}
\label{table3}
{\small
\begin{tabular}{l|ccc|ccc}
\hline
\textbf{Datasets} &
  \multicolumn{3}{c|}{COCO2017} &
  \multicolumn{3}{c}{Flickr30K}  \\ \hline
\textbf{Evaluation Metric} &
  R@5 \textcolor{red}{$\uparrow$} &
  R@10 \textcolor{red}{$\uparrow$} &
  MS@1000 (G$\times$ST) \textcolor{blue}{$\downarrow$} &
  R@5 \textcolor{red}{$\uparrow$} &
  R@10 \textcolor{red}{$\uparrow$} &
  MS@1000 (G) \textcolor{blue}{$\downarrow$} \\ \hline

\textbf{Baseline CLIP (ResNet50)} &
71.4$\pm$0.5 & 81.8$\pm$0.1 & 13.3$\pm$0.3 &
78.5$\pm$0.6 & 85.9$\pm$0.4 & 18.6$\pm$0.5 \\ \hline

\textbf{\faCheck $\,$ \textcolor{orange}{\faFire} \textcolor{red}{I\&T} SFID} &
[66.6]$\pm$0.2 & 76.3$\pm$0.3 & 12.7$\pm$0.4 &
74.9$\pm$0.6 & 79.6$\pm$0.2 & 15.2$\pm$0.3 \\ \hline

\textbf{\faCheck $\,$ \textcolor{orange}{\faFire} \textcolor{red}{I\&T} PromptArray} &
66.5$\pm$0.5 & [76.5]$\pm$0.4 & 12.2$\pm$0.1 &
74.9$\pm$0.3 & [81.6]$\pm$0.3 & 15.4$\pm$0.5 \\ \hline

\textbf{\faCheck $\,$ \textcolor{orange}{\faFire} \textcolor{red}{I\&T} FairerCLIP} &
65.8$\pm$0.6 & 74.8$\pm$0.3 & 11.8$\pm$0.5 &
75.2$\pm$0.2 & [81.6]$\pm$0.4 & 16.0$\pm$0.3 \\ \hline

\textbf{\faCheck $\,$ \textcolor{cyan}{\faSnowflake} \textcolor{red}{I\&T} CLIP-clip} &
65.6$\pm$0.1 & 74.9$\pm$0.5 & \textbf{[10.0]}$\pm$0.6 &
[75.3]$\pm$0.3 & 80.8$\pm$0.4 & \textbf{[13.2]}$\pm$0.5 \\ \hline

\textbf{\faTimes $\,$ \textcolor{cyan}{\faSnowflake} \textcolor{red}{T} Orth-Proj} &
64.2$\pm$0.2 & 72.5$\pm$0.6 & 11.2$\pm$0.4 &
74.9$\pm$0.5 & 78.7$\pm$0.2 & 14.1$\pm$0.6 \\ \hline

\textbf{\faTimes $\,$ \textcolor{cyan}{\faSnowflake} \textcolor{red}{T} Orth-Cali} &
64.7$\pm$0.4 & 74.6$\pm$0.3 & 11.0$\pm$0.5 &
74.9$\pm$0.2 & 80.4$\pm$0.6 & 16.1$\pm$0.3 \\ \hline

\rowcolor{gray!30}
\textbf{\faTimes $\,$ \textcolor{cyan}{\faSnowflake} \textcolor{red}{I\&T} Ours} &
\textbf{[68.3]}$\pm$0.5 & \textbf{[78.4]}$\pm$0.2 & [10.1]$\pm$0.6 &
\textbf{[76.6]}$\pm$0.4 & \textbf{[85.1]}$\pm$0.3 & [13.6]$\pm$0.1 \\ \hline

\end{tabular}%
}
\end{table*}

\begin{table*}[h] \centering
\renewcommand{\arraystretch}{1.3}
\setlength{\tabcolsep}{2pt}
\caption{Experimental results of BLIP on the zero-shot image classification task.}
\label{table2}

{\small
\begin{tabular}{l|c|cc|c|cc}
\hline
\textbf{Datasets} & \multicolumn{3}{c|}{CelebA} & \multicolumn{3}{c}{FACET} \\
\hline
\textbf{Evaluation Metric} &
F1 \textcolor{red}{$\uparrow$} &
$\Delta_{\text{EO}}^{\text{Avg}}$ (G$\times$A) \textcolor{blue}{$\downarrow$} &
$\Delta_{\text{EO}}^{\text{Max}}$ (G$\times$A) \textcolor{blue}{$\downarrow$} &
Macro F1 \textcolor{red}{$\uparrow$} &
$\Delta_{\text{EO}}^{\text{Avg}}$ (G) \textcolor{blue}{$\downarrow$} &
$\Delta_{\text{EO}}^{\text{Max}}$ (G) \textcolor{blue}{$\downarrow$} \\
\hline

\textbf{Baseline BLIP} &
46.9$\pm$0.6 & 14.7$\pm$0.4 & 29.2$\pm$0.6 & 68.4$\pm$0.3 & 8.8$\pm$0.5 & 48.0$\pm$0.4 \\ \hline

\textbf{\faCheck $\,$ \textcolor{orange}{\faFire} \textcolor{red}{I\&T} SFID} &
[52.5]$\pm$0.2 & \textbf{[12.1]}$\pm$0.1 & 26.4$\pm$0.3 & 60.6$\pm$0.1 & 8.2$\pm$0.2 & 47.1$\pm$0.5 \\ \hline

\textbf{\faCheck $\,$ \textcolor{orange}{\faFire} \textcolor{red}{I\&T} FairerCLIP} &
52.1$\pm$0.4 & 13.4$\pm$0.6 & \textbf{[22.6]}$\pm$0.4 & [61.0]$\pm$0.3 & 8.9$\pm$0.6 & 46.5$\pm$0.3 \\ \hline

\textbf{\faTimes $\,$ \textcolor{orange}{\faFire} \textcolor{red}{I\&T} PRISM} &
[5]$\pm$0.5 & 13.8$\pm$0.2 & 26.1$\pm$0.5 & 60.3$\pm$0.4 & \textbf{[6.8]}$\pm$0.3 & 46.0$\pm$0.5 \\ \hline

\textbf{\faTimes $\,$ \textcolor{cyan}{\faSnowflake} \textcolor{red}{I\&T} PRISM-mini} &
49.5$\pm$0.4 & 14.0$\pm$0.5 & 25.1$\pm$0.1 & 60.0$\pm$0.2 & 7.4$\pm$0.1 & 46.1$\pm$0.4 \\ \hline

\textbf{\faTimes $\,$ \textcolor{cyan}{\faSnowflake} \textcolor{red}{I\&T} RoboShot} &
52.1$\pm$0.1 & 14.7$\pm$0.1 & 25.1$\pm$0.3 & 60.3$\pm$0.5 & 7.6$\pm$0.6 & \textbf{[44.4]}$\pm$0.1 \\ \hline

\textbf{\faTimes $\,$ \textcolor{cyan}{\faSnowflake} \textcolor{red}{T} Orth-Proj} &
48.2$\pm$0.3 & 14.6$\pm$0.2 & 27.0$\pm$0.2 & 58.3$\pm$0.2 & 8.5$\pm$0.2 & 46.3$\pm$0.2 \\ \hline

\textbf{\faTimes $\,$ \textcolor{cyan}{\faSnowflake} \textcolor{red}{T} Orth-Cali} &
49.6$\pm$0.6 & 14.4$\pm$0.1 & 27.5$\pm$0.6 & 59.5$\pm$0.3 & 8.0$\pm$0.6 & 46.0$\pm$0.5 \\ \hline

\rowcolor{gray!30}
\textbf{\faTimes $\,$ \textcolor{cyan}{\faSnowflake} \textcolor{red}{I\&T} Ours} &
\textbf{[54.8]}$\pm$0.4 & [12.2]$\pm$0.5 & [23.2]$\pm$0.3 & \textbf{[68.4]}$\pm$0.3 & [7.3]$\pm$0.4 & [44.7]$\pm$0.3 \\ \hline

\end{tabular}
}
\end{table*}

\clearpage

\begin{table*}[h]
\centering
\renewcommand{\arraystretch}{1.1}
\setlength{\tabcolsep}{4pt}
\caption{Experimental results of BLIP on the text-to-image retrieval task.}
\label{table3}
{\small
\begin{tabular}{l|ccc|ccc}
\hline
\textbf{Datasets} &
  \multicolumn{3}{c|}{COCO2017} &
  \multicolumn{3}{c}{Flickr30K}  \\ \hline

\textbf{Evaluation Metric} &
  R@5 \textcolor{red}{$\uparrow$} &
  R@10 \textcolor{red}{$\uparrow$} &
  MS@1000 (G$\times$ST) \textcolor{blue}{$\downarrow$} &
  R@5 \textcolor{red}{$\uparrow$} &
  R@10 \textcolor{red}{$\uparrow$} &
  MS@1000 (G) \textcolor{blue}{$\downarrow$} \\ \hline

\textbf{Baseline BLIP} &
93.9$\pm$0.5 & 97.2$\pm$0.1 & 15.1$\pm$0.6 &
93.9$\pm$0.4 & 96.8$\pm$0.2 & 18.3$\pm$0.6 \\ \hline

\textbf{\faCheck $\,$ \textcolor{orange}{\faFire} \textcolor{red}{I\&T} SFID} &
[90.6]$\pm$0.3 & 95.9$\pm$0.4 & 12.5$\pm$0.1 &
86.1$\pm$0.6 & 89.9$\pm$0.5 & 13.5$\pm$0.1 \\ \hline

\textbf{\faCheck $\,$ \textcolor{orange}{\faFire} \textcolor{red}{I\&T} PromptArray} &
90.4$\pm$0.2 & [96.0]$\pm$0.3 & 12.9$\pm$0.4 &
86.4$\pm$0.3 & 91.5$\pm$0.4 & 12.8$\pm$0.5 \\ \hline

\textbf{\faCheck $\,$ \textcolor{orange}{\faFire} \textcolor{red}{I\&T} FairerCLIP} &
90.0$\pm$0.6 & 94.7$\pm$0.2 & 13.3$\pm$0.3 &
[91.0]$\pm$0.2 & [91.7]$\pm$0.3 & 12.0$\pm$0.4 \\ \hline

\textbf{\faCheck $\,$ \textcolor{cyan}{\faSnowflake} \textcolor{red}{I\&T} CLIP-clip} &
90.3$\pm$0.4 & 94.5$\pm$0.6 & \textbf{[11.2]}$\pm$0.5 &
[91.0]$\pm$0.1 & 90.8$\pm$0.1 & \textbf{[8.1]}$\pm$0.2 \\ \hline

\textbf{\faTimes $\,$ \textcolor{cyan}{\faSnowflake} \textcolor{red}{T} Orth-Proj} &
90.8$\pm$0.5 & 92.8$\pm$0.4 & 13.5$\pm$0.2 &
90.7$\pm$0.6 & 88.1$\pm$0.6 & 8.8$\pm$0.3 \\ \hline

\textbf{\faTimes $\,$ \textcolor{cyan}{\faSnowflake} \textcolor{red}{T} Orth-Cali} &
90.3$\pm$0.1 & 94.1$\pm$0.5 & 13.7$\pm$0.5 &
90.7$\pm$0.3 & 89.9$\pm$0.3 & 9.0$\pm$0.1 \\ \hline

\rowcolor{gray!30}
\textbf{\faTimes $\,$ \textcolor{cyan}{\faSnowflake} \textcolor{red}{I\&T} Ours} &
\textbf{[92.2]}$\pm$0.4 & \textbf{[96.7]}$\pm$0.2 & [12.0]$\pm$0.3 &
\textbf{[93.3]}$\pm$0.5 & \textbf{[96.5]}$\pm$0.6 & [8.4]$\pm$0.5 \\ \hline

\end{tabular}%
}
\end{table*}

\clearpage

\noindent\textbf{\Large \textcolor{red}{G}. More Illustrative Examples}\\

\begin{figure*}[h]
  \centering
  \includegraphics[height=19cm, width=17.4cm]{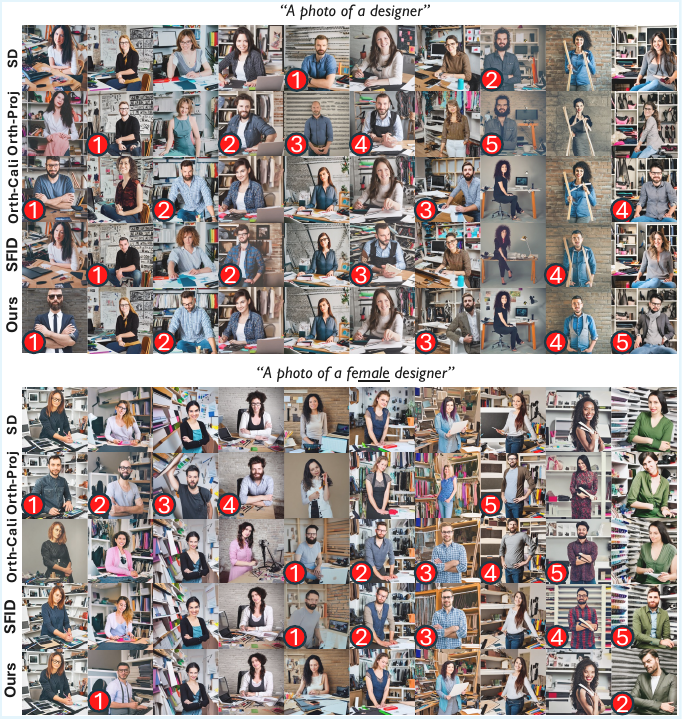}
\caption{Illustrative examples for $o=\text{``designer"}$. We randomly sample ten generated images for each method. Male-looking samples are marked in \textcolor{red}{red} and numbered. On the left, a more balanced ratio of female- and male-looking samples indicates lower bias, while on the right, fewer male-looking samples reflect better preservation of self-utility.}
  \label{fig4}
\end{figure*}

\begin{figure*}[h]
  \centering
  \includegraphics[height=19cm, width=17.4cm]{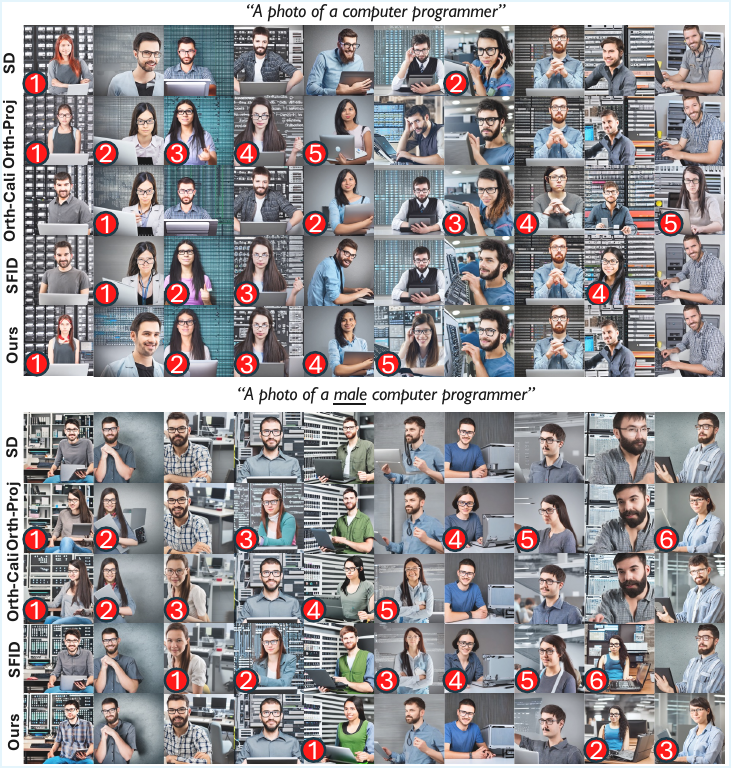}
\caption{Illustrative examples for $o=\text{``computer programmer"}$. We randomly sample ten generated images for each method. Female-looking samples are marked in \textcolor{red}{red} and numbered. On the left, a more balanced ratio of female- and male-looking samples indicates lower bias, while on the right, fewer female-looking samples reflect better preservation of self-utility.}
  \label{fig4}
\end{figure*}

\begin{figure*}[h]
  \centering
  \includegraphics[height=19cm, width=17.4cm]{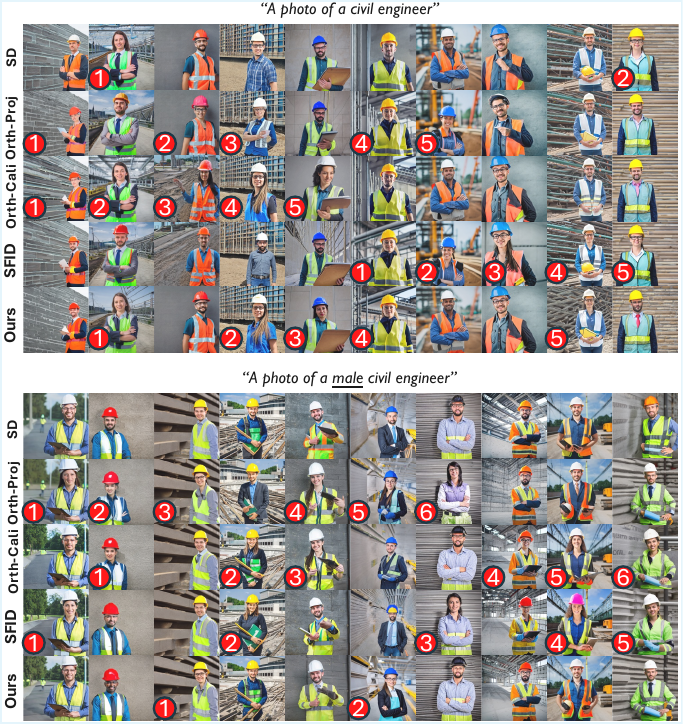}
\caption{Illustrative examples for $o=\text{``civil engineer"}$. We randomly sample ten generated images for each method. Female-looking samples are marked in \textcolor{red}{red} and numbered. On the left, a more balanced ratio of female- and male-looking samples indicates lower bias, while on the right, fewer female-looking samples reflect better preservation of self-utility.}
  \label{fig4}
\end{figure*}

\begin{figure*}[h]
  \centering
  \includegraphics[height=19cm, width=17.4cm]{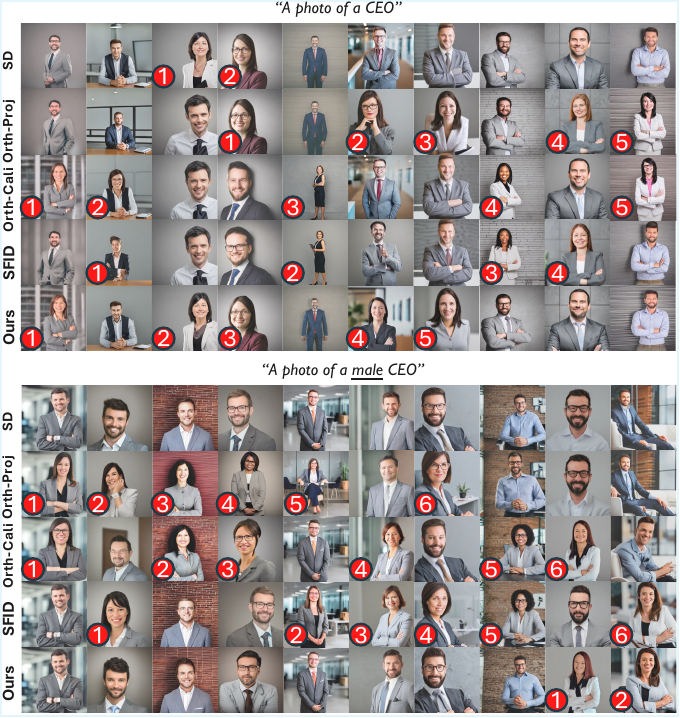}
\caption{Illustrative examples for $o=\text{``CEO"}$. We randomly sample ten generated images for each method. Female-looking samples are marked in \textcolor{red}{red} and numbered. On the left, a more balanced ratio of female- and male-looking samples indicates lower bias, while on the right, fewer female-looking samples reflect better preservation of self-utility.}
  \label{fig4}
\end{figure*}

\clearpage

%  {
%     \small
%     \bibliographystyle{ieeenat_fullname}
%     \bibliography{main}
% }

\end{document}